\documentclass[letterpaper, 10 pt, conference]{ieeeconf}  % Comment this line out if you need a4paper
\IEEEoverridecommandlockouts                              % This command is only needed if
\usepackage[utf8]{inputenc} % Input encoding
\usepackage[T1]{fontenc} % Font (output) encoding, required to copy link texts
\usepackage{cite} % For references, required to combine references
\usepackage{graphics} % For pdf, bit-mapped graphics files
\usepackage{times} % Assumes new font selection scheme installed
\usepackage{amsmath} % Assumes amsmath package installed
\usepackage{amssymb} % Assumes amsmath package installed
\usepackage{amsfonts}
\usepackage{mathtools} % Load the mathtools package for \mathclap
\usepackage{graphicx}
\usepackage{textcomp}
\usepackage{tikz}
\usepackage{siunitx} % For SI units
\usepackage{array}
\newcolumntype{P}{>{$}r<{$}@{\,\ensuremath{\pm}\,}>{$}l<{$}}
\usepackage{wrapfig}
\usepackage{tabularx}
\usepackage{booktabs} % Lines for tables
\usepackage{makecell}
\usepackage{multirow} % Multiple rows per column in tables
\usepackage{hhline}

\usepackage{tikz}
\usetikzlibrary{backgrounds, fit, shapes.geometric, positioning}
\usepackage{url}
\usepackage{xcolor}
\usepackage{soul}
\usepackage{rotating}

\usepackage{dsfont}

\usepackage[font=footnotesize]{caption} % Set table caption below Table for longer descriptions
\usepackage[font=footnotesize, labelfont=footnotesize, textfont=footnotesize, labelformat=brace]{subcaption} % Sub-captions
\makeatletter % Needed for @ in \p@subfigure, see https://tex.stackexchange.com/questions/8351/what-do-makeatletter-and-makeatother-do
\renewcommand\p@subfigure{\thefigure\,}
\makeatother % Needed for @ in \p@subfigure

\usepackage[hidelinks,bookmarksnumbered,bookmarksdepth=1]{hyperref} % Links etc.
\usepackage[ruled, linesnumbered, vlined, noend]{algorithm2e}

\SetCommentSty{mycommfont}
\usepackage[noend]{algpseudocode}

\usepackage{balance} % Balance two columns
\def\secref#1{Sec.~\ref{#1}}
\def\figref#1{Fig.~\ref{#1}}

\def\tabref#1{Tab.~\ref{#1}}

\def\eqref#1{Eq.~(\ref{#1})}

\algnewcommand{\algorithmicgoto}{\textbf{go to}}
\algnewcommand{\Goto}[1]{\algorithmicgoto~line~\ref{#1}}

\newcommand\etal{~\emph{et al.}}

\newcolumntype{Y}{>{\centering\arraybackslash}X}% New Y, equal sized as X but centered
\title{\LARGE \bf Interactive Shaping of Granular Media Using Reinforcement Learning}

\author{Benedikt Kreis$^{1,3,4}$ \and Malte Mosbach$^{2,4}$ \and Anny Ripke$^{1}$ \and M. Ehsan Ullah$^{1}$ \and Sven Behnke$^{2,3,4}$ \and Maren Bennewitz$^{1,3,4}$% <-this % stops a space
  \thanks{
  	\hspace{-1.05em}$^{1}$Humanoid Robots Lab, University of Bonn, Germany.\newline
        $^{2}$Autonomous Intelligent Systems Lab, University of Bonn, Germany.\newline
  	$^{3}$Center for Robotics, University of Bonn, Germany.\newline
  	$^{4}$Lamarr Institute for ML and AI, Bonn, Germany.\newline
         This work has been partially funded by the EC, grant No. 964854 RePAIR H2020-FETOPEN-2018-2020 and by the BMBF within the Robotics Institute Germany, grant No. 16ME0999.\newline
	The corresponding author is Benedikt Kreis: \href{mailto:kreis@cs.uni-bonn.de}{kreis@cs.uni-bonn.de}}%
}
\begin{document}
\maketitle
\thispagestyle{empty} 
\pagestyle{empty}

%%%%%%%%%%%%%%%%%%%%%%%%%%%%%%%%%%%%%%%%%%%%%%%%%%%%%%%%%%%%%%%%%%%%%%%%%%%%%%%%
\begin{abstract}
Autonomous manipulation of granular media, such as sand, is crucial for applications in construction, excavation, and additive manufacturing.
However, shaping granular materials presents unique challenges due to their high-dimensional configuration space and complex dynamics, where traditional rule-based approaches struggle without extensive engineering efforts.
Reinforcement learning (RL) offers a promising alternative by enabling agents to learn adaptive manipulation strategies through trial and error.
In this work, we present an RL framework that enables a robotic arm with a cubic end-effector and a stereo camera to shape granular media into desired target structures.
We show the importance of compact observations and concise reward formulations for the large configuration space, validating our design choices with an ablation study.
Our results demonstrate the effectiveness of the proposed approach for the training of visual policies that manipulate granular media including their real-world deployment, significantly outperforming two baseline approaches in terms of target shape accuracy.
\end{abstract}

%%%%%%%%%%%%%%%%%%%%%%%%%%%%%%%%%%%%%%%%%%%%%%%%%%%%%%%%%%%%%%%%%%%%%%%%%%%%%%%%
\section{Introduction}
\label{sec:intro}
The ability to manipulate granular media such as sand has many applications in robotics, ranging from construction and excavation~\cite{schenckLearningRoboticManipulation2017, jinLearningExcavationRigid2023, zhuFewshotAdaptationManipulating2023, judRoboticEmbankmentFreeform2021, egliSoilAdaptiveExcavationUsing2022, osaDeepReinforcementLearning2022, kurinovAutomatedExcavatorBased2020, luExcavationReinforcementLearning2022}
to additive manufacturing~\cite{cherubiniModelfreeVisionbasedShaping2020}.
Unlike the manipulation of rigid bodies, the shaping of granular media is accompanied by unique challenges due to their particle nature.
Accurate modeling and control of such media, requires accounting for complex interactions that may vary depending on the material composition.
To successfully shape such media, an agent must continuously adapt its manipulation strategy in response to material deformation.
Applying traditional modeling approaches requires extensive engineering efforts due to the large configuration space of deformable objects and media~\cite{matasSimtoRealReinforcementLearning2018}.

\begin{figure}[t]
    \centering
    \includegraphics[width=0.94\linewidth]{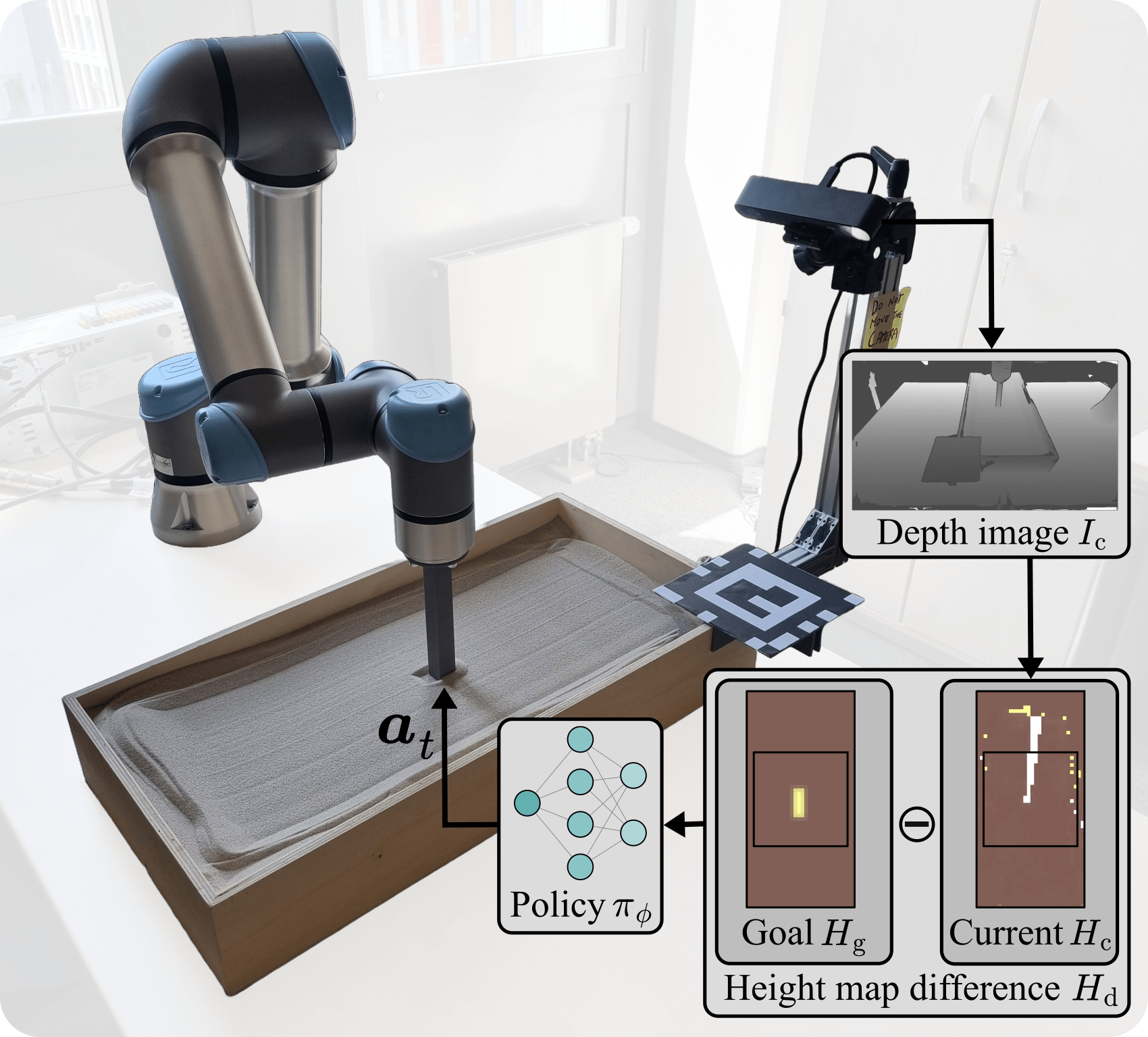}
    \caption{The robot's task is to manipulate the granular media with its cubic end-effector to shape it as close as possible to desired goal configurations.
    The configurations are abstracted as height maps and the robot reconstructs the height map corresponding to the current configuration from depth observations.
    Our approach closes the loop between partial visual observations and goal-oriented manipulation, accounting for the dynamics of collapsing granular media during the manipulation.
    }
    \label{teaser}
\end{figure}

Reinforcement learning (RL) provides an alternative forgoing the need to predict the precise consequences of manipulations to the granular media, allowing the agent instead to adaptively react to the way the manipulation unfolds by learning optimal strategies through trial and error.
This allows the robot to interact with the state of the medium in a closed loop (see \figref{teaser}).
Although RL has been successfully applied to the dexterous manipulation of rigid
objects~\cite{chenSystemGeneralInHand2022a}, recent research suggests its potential for manipulating deformable objects, including cloth handling~\cite{matasSimtoRealReinforcementLearning2018} and fluid control~\cite{fontDeepReinforcementLearning2025}.
However, the application of RL to the interactive manipulation of granular media has not been explored sufficiently, likely due to two key challenges.
First, \emph{finding a compact observation space} for granular media is difficult.
Rigid objects can often be represented efficiently using their poses, but granular media exhibits an effectively infinite configuration space, requiring a high-dimensional representation.
Second, \emph{designing an effective reward function} is challenging.
While object manipulation tasks often leverage distance-based rewards to guide learning, shaping rewards this way for granular media results in very sparse rewards since most random manipulation actions lead to configurations further away from the desired goal, which can result in an agent that avoids interactions with the medium.

In this work, we address these challenges by studying how the Markov decision process (MDP) of granular media manipulation can be defined to make RL algorithms sucessfully applicable.
To this end, we make the following contributions:
\begin{itemize}
    \item We develop a novel reward formulation that fosters fast and stable convergence of RL training towards functional granular media manipulation behaviors.
    \item We demonstrate that RL policies can be learned from visual observations by converting high-dimensional depth images to compact height map representations.
    \item We demonstrate that the resulting formulation allows for zero-shot transfer of trained policies to a real robot.
\end{itemize}

%%%%%%%%%%%%%%%%%%%%%%%%%%%%%%%%%%%%%%%%%%%%%%%%%%%%%%%%%%%%%%%%%%%%%%%%%%%%%%%%
\section{Related Work}
\label{sec:related}
Manipulating granular media like sand is an active research field in robotics tackling challenges in simulating contacts for locomotion~\cite{zhuDatadrivenApproachFast2019, karsaiRealTimeRemodelingGranular2022, kerimogluLearningManipulationSteep2024}, retrieving objects~\cite{xuTactilebasedObjectRetrieval2024}, grading~\cite{mironAutonomousGradingReal2022}, sweeping~\cite{alaturMaterialAgnosticShapingGranular2023, xueNeuralFieldDynamics2023}, trenching~\cite{pavlovSoilDisplacementTerramechanics2019}, and excavating~\mbox{\cite{schenckLearningRoboticManipulation2017, jinLearningExcavationRigid2023, zhuFewshotAdaptationManipulating2023, judRoboticEmbankmentFreeform2021, egliSoilAdaptiveExcavationUsing2022, osaDeepReinforcementLearning2022, kurinovAutomatedExcavatorBased2020, luExcavationReinforcementLearning2022, liuLocalizedGraphBasedNeural2025}}.
The scale of the proposed pipelines varies from sweeping a limited amount of grit stones~\cite{alaturMaterialAgnosticShapingGranular2023} to shaping entire landscapes~\cite{hurkxkensRoboticLandscapesDesigning2020}.

\subsection{Simulating Granular Media}
\label{subsec:sand_sim}
Modeling contact interactions between robots and granular media is challenging due to the amount of individual, simulated particles \cite{tuomainenManipulationGranularMaterials2022}.
Exact methods like finite element methods (FEMs) or discrete element methods (DEMs) are suitable to simulate material deformations and physically accurate behavior of granular media.
While FEM is less computationally expensive than DEM, it does not capture the granularity of particle interactions, as it treats materials as continua.
Although both simulate a larger variety of phenomena than rigid body simulations, their required extensive computations can hinder their application in robotics~\cite{zhuDatadrivenApproachFast2019}.
Therefore, Xu\etal~\cite{xuTactilebasedObjectRetrieval2024} use relatively big and coarse particles to reduce their total amount.
This minimizes the computational cost, but it makes the model less accurate.

Another way to save computational cost is to model particles as a large graph and selectively activate small subgraphs to predict how local robot-terrain interactions deform the granular medium, as proposed by Liu\etal~\cite{liuLocalizedGraphBasedNeural2025}.

Kim\etal~\cite{kimDevelopingSimpleModel2019} as well as Pavlov and Johnson~\cite{pavlovSoilDisplacementTerramechanics2019} go one step further by focusing on deformations on the surface of granular media, rather than computing the behavior of each particle.
They proposed a sand model using a height map that mimics the collapse of sand piles based on the angle of repose.
If the angle of repose between two height map cells is surpassed, the excess sand is distributed to adjacent cells until the angle condition is met again.
As their model is both computationally efficient and physically realistic, we base our granular media simulation on the same model.
\begin{figure*}[t] 	
    \centering
    \includegraphics[width=\linewidth]{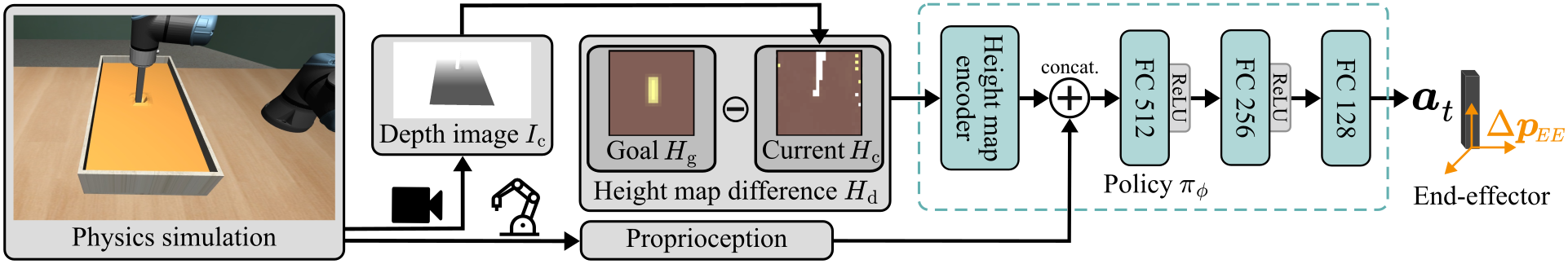}
    \caption{
        Overview of our approach: 
        We employ a training process to enable agents to manipulate granular media using sensory inputs.
        We train a visual policy via reinforcement learning to realize goal shape configurations using the difference between the current and the desired goal height map.
        The current height map is reconstructed from depth images.
        The height map difference is fed into our height map encoder and concatenated with the robot's proprioception.
        The resulting policy controls the end-effector to shape the granular media.
        }
    \label{fig:architecture}
\end{figure*}

\subsection{Robotic Manipulation of Granular Media}
\label{subsec:sand_shaping}
A common construction task is the grading of sand, e.g., to build roads. Miron\etal~\cite{mironAutonomousGradingReal2022} leverage imitation learning to control a bulldozer to level sand piles modeled as two-dimensional multivariate Gaussian distributions.

Instead of leveling piles, Alatur\etal~\cite{alaturMaterialAgnosticShapingGranular2023} proposed to form them by teaching a robot to sweep grit stones and wooden chips on a table.
By extending a classical motion planner with optimal transport, their robot is able to arrange predetermined pile shapes.

Another well-researched construction task is excavating granular media. Schenck\etal~\cite{schenckLearningRoboticManipulation2017} learned predictive models with highly-tailored convolutional neural network (CNN) architectures based on an experimental data set.
This allows to predict the dynamics of scooping and dumping actions.
Jin\etal~\cite{jinLearningExcavationRigid2023} tackled the excavation task with the offline RL algorithm Implicit Q-Learning (IQL) and trained on a prerecorded data set with six different terrain types.
IQL allows them to outperform sub-optimal demonstrations in the data set, but it requires new data to adapt the policy to new terrain types.

Several authors have used an autonomous walking excavator called HEAP~\cite{judHEAPAutonomousWalking2021}.
Using this heavy machinery, they proposed multiple approaches ranging from classical path planning focusing on the overall landscaping system~\cite{judRoboticEmbankmentFreeform2021, hurkxkensRoboticLandscapesDesigning2020, terenziAutonomousExcavationPlanning2024} to an RL-based approach~\cite{egliSoilAdaptiveExcavationUsing2022}.
In the latter, they use Proximal Policy Optimization (PPO) with general advantage estimation (GAE) to train a controller to adaptively dig in different soil types.
Further RL-based excavation approaches for heavy machinery have been proposed by Kurinov\etal~\cite{kurinovAutomatedExcavatorBased2020} using Proximal Policy Optimization with Covariance Matrix Adaptation (PPO-CMA), as well as by Osa and Aizawa~\cite{osaDeepReinforcementLearning2022} using Qt-Opt, a sample efficient variant of Q-learning trained from depth images.

For extraterrestrial missions, being able to reuse existing equipment is a crucial advantage due to the saved weight.
Therefore, Pavlov and Johnson~\cite{pavlovSoilDisplacementTerramechanics2019} proposed the idea to dig trenches with rover wheels instead of a dedicated end-effector.
Kim\etal~\cite{kimDevelopingSimpleModel2019} built upon Pavlov's experimental results and they proposed three methods to dig trenches of constant depths.
Among them, a classical \(A^*\)-based planner performing single strokes and a learned approach based on Deep Deterministic Policy Gradient (DDPG) combined with Hindsight Experience Replay (HER) performing multiple strokes. 

Apart from weight restrictions, extraterrestrial systems can suffer domain-shift issues due to partially known environmental conditions when training on earth and then deploying to outer space.
In particular, vision-based systems struggle with unforeseen conditions.
To overcome this problem, Zhu\etal~\cite{zhuFewshotAdaptationManipulating2023} proposed an adaptive, vision-based scooping strategy leveraging meta-learning on a deep Gaussian process and show that despite the domain shift, depth images allow to learn the scooping task.

Yet another challenge is to move the robot itself within granular media, especially on steep, inclined surfaces.
That is why, Karsai\etal~\cite{karsaiRealTimeRemodelingGranular2022} and Kerimoglu\etal~\cite{kerimogluLearningManipulationSteep2024} explored gaits to manipulate local granular terrain to improve the climbing and turning performance of a wheeled-legged robot using Bayesian optimization.
In addition to direct terrain manipulation, Hu\etal~\cite{huLearningGranularMedia2025} proposed to indirectly manipulate objects by creating small avalanches within granular media.
They actuated a robot's leg to trigger the collapse of piles, while a vision transformer is trained to capture the avalanche dynamics.

Using Kim\etal's~\cite{kimDevelopingSimpleModel2019} efficient sand model allows us to train an RL agent in an online manner overcoming shortcomings of suboptimal demonstrations in the data set as in~\cite{jinLearningExcavationRigid2023}.
We adopt the idea of using height maps as observations, but instead of a stroke-based approach, where end-effector motions are limited to straight lines between two \(x\)-\(y\) coordinates at a fixed \(z\) height in the granular medium, we let the agent freely decide which motions to take in any Cartesian direction at each step.
Also, we do not limit the agent to binary heights so that our agent can shape structures of varying depth.
To perceive the state of the granular media, we make use of depth images like~\cite{osaDeepReinforcementLearning2022, zhuFewshotAdaptationManipulating2023}, but we extract the relevant state information by representing them as height maps, which is computationally more efficient.

%%%%%%%%%%%%%%%%%%%%%%%%%%%%%%%%%%%%%%%%%%%%%%%%%%%%%%%%%%%%%%%%%%%%%%%%%%%%%%%%
\section{Method}
\label{sec:main}
In this section, we present our RL framework that learns how to move a robotic end-effector (EE) in granular media to create diverse goal shapes.

%%%%%%%%%%%%%%%%%%%%%%%%
\subsection{Overview}
\label{subsec:overview}
\figref{fig:architecture} provides an overview of our learning framework.
Using observations from a physics simulation, we train a visual policy to manipulate granular media using the difference between the reconstructed current height map and the height map representing the desired goal configuration.

%%%%%%%%%%%%%%%%%%%%%%%%
\subsection{Task Description}
\label{subsec:task}
The goal of the agent during one episode is to form the granular media according to a physically viable goal shape.
We focus on shapes that deepen an initially flat surface.
The shape is given in the form of a height map as shown in \figref{fig:architecture}.
When a training episode starts the robot is initialized to a random start configuration with the EE laying within a virtual cuboid of \(30\)\,\texttimes\,\(30\)\,\texttimes\,\SI{5}{\cubic\centi\meter}, which is located \SI{2}{cm} above the granular media. The granular media is initialized as a flat bed at a height of \SI{6}{cm}.
From the starting position, the agent has to move the EE through the media to shape it to resemble the goal height map as much as possible.
An episode terminates after a fixed number of time steps \(N_\text{ep}\).
At the end of an episode, we reset the environment and the agent receives a new goal height map.

%%%%%%%%%%%%%%%%%%%%%%%%
\subsection{Architecture}
\label{subsec:architecture}
The underlying idea of RL is to model and optimize transitions from one state to the next \(s_t \rightarrow s_{t+1}\) as an MDP.
In this process, the reward \(r_t = r(s_t, a_t)\) incentivizes the RL agent to take actions \(a_t = \pi_{\phi}(s_t)\) at the time step \(t\) with respect to a policy \(\pi_{\phi}\).
Commonly, pairs of tuples \(\left(s_t, a_t, r_t, s_{t+1}\right)\) summarize states and actions.
The agent's goal is to maximize the cumulative return \(R = \sum^{T}_{i=t} \gamma^{(i-t)}r_t\) of the \(\gamma\)-discounted rewards.

\subsubsection*{\bfseries{RL Algorithm}} 
\label{subsubsec:rl_algo}
In this work, we use the off-policy algorithm Truncated Quantile Critics (TQC)~\cite{kuznetsovControllingOverestimationBias2020} with two critics and \(25\) quantiles, which showed the best training convergence among three tested RL algorithms~\mbox{(see \secref{subsec:results})}.
To assure the agent's exploration, we add Gaussian noise from a process \(\mathcal{N}\) with a standard deviation~\(\sigma_{\epsilon_\pi}\) to the actions, so that \(a_t = \pi_{\phi}(s_t) + \mathcal{N}(0, \sigma_{\epsilon_\pi})\).

\subsubsection*{\bfseries{Action Space}} 
\label{subsubsec:action_space}
Per time step, our agent moves the robot's EE in continuous action increments \((\Delta x,\Delta y,\Delta z)\) using an operational space controller (OSC)~\cite{khatibUnifiedApproachMotion1987}.
At each step the agent can move the EE in increments of up to \SI{4}{cm} in each direction or it can keep it still at the current position.
We normalize all actions to values in the interval \([-1,+1]\).

\subsubsection*{\bfseries{Observation Space}} 
\label{subsubsec:observation_space}
\begin{table}[b]
	\centering
    	\begingroup
	\caption{Observation space.}
	\label{tab:observations}
        \begin{tabularx}{\linewidth}{lYcc}
		\toprule[\lightrulewidth]
		\multicolumn{1}{c}{Observation} & \multicolumn{1}{c}{Notation}& \multicolumn{1}{c}{Size} \\ 
		\midrule[\lightrulewidth]
            Current EE position & \((x_c,y_c,z_c)_\text{EE}\) & 3 \\
            Previous EE position & \((x_p,y_p,z_p)_\text{EE}\) & 3 \\
  		Difference height map & \(H_\text{d}\) & \multirow{3}{*}{\Bigg\}}\,\multirow{3}{*}{64} \\
            EE mask & \(M_\text{EE}\) & \\
            Goal mask & \(M_\text{g}\)  & \\
		\bottomrule[\lightrulewidth]
		& \multicolumn{1}{r}{Sum} & 70 \\
	\end{tabularx}
    	\endgroup
\end{table}

The task-relevant information included in the observation space is shown in~\tabref{tab:observations}.
We normalize all observations to values in the interval~\([-1,+1]\).
For the EE, we include its current \((x_c,y_c,z_c)_\text{EE}\) and previous position \((x_p,y_p,z_p)_\text{EE}\) so that the agent can adequately scale the inputs for the OSC.
Furthermore, we use the difference height map \(H_\text{d} = H_\text{g} - H_\text{c}\), obtained from the goal \(H_\text{g}\) and the current height map \(H_\text{c}\).
During the reconstruction process of the current height map, we convert the current depth image \(I_\text{c}\) into 3D points and fit a grid on top of the ones within the granular media.
The mean of all points that fall into one grid cell define the resulting cell elevation.
To indicate the current EE position and the goal area in the same compact representation as the current height map, we create two Boolean masks of the same shape.
The true values in the EE mask \(M_\text{EE}\) represent the two-dimensional projection to the \(xy\)-plane of the EE's base shape, while the true values in the goal mask \(M_\text{g}\) represent the grid cells that are unequal to the initialization height of the goal height map.
To reduce the observation size, we combine the difference height map with the two masks using our height map encoder (see~\figref{fig:architecture}).
It ensures that the information of the EE's current position and the goal area is available in the same space as the height map elevation data.
To achieve that, we first stack the difference height map with the EE mask.
Then we extract feature vectors of size 64 from both, the stacked observation and the goal mask using a CNN and amplify the stacked features within the goal area by multiplying them with the sigmoid function of the goal mask, which works like a gating mechanism.
The resulting feature vector of size 64 is concatenated with a vector of all other EE observations.
The height map encoder is designed for inputs of size \(32\)\,\texttimes\,\(32\).
Depending on the available manipulation area and goal area size, we reduce the input size by padding the outer cells with zero.
In our implementation, all height maps and masks have a grid cell size of \(1\)\,\texttimes\,\(\SI{1}{\square\centi\meter}\)
and we limit the height map elevation values to~\([\SI{0}{cm},\SI{20}{cm}]\).

%%%%%%%%%%%%%%%%%%%%%%%%
\subsection{Reward Function}
\label{subsec:reward}
\begin{figure}[t]
    \centering
    \includegraphics[width=\linewidth]{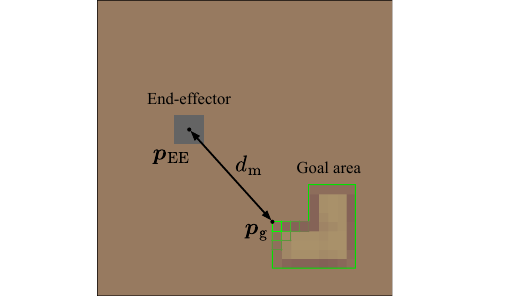}
    \caption{We employ the distance \(d_\text{m}\) of the EE position \(\boldsymbol{p}_\text{EE}\) to the closest point belonging to the goal area \(\boldsymbol{p}_\text{g}\) to guide the agent towards regions that require manipulation through a dense reward signal (see \secref{subsec:reward}).}
    \label{fig:reward_shaping}
\end{figure}
Unlike rigid-body tasks, where distances to goal positions are well-defined and straightforward to compute, the high-dimensional configuration space of granular materials make reward shaping significantly more challenging.
A random manipulation is often more likely to increase the distance to the goal configuration than to reduce it, resulting in an unbalanced reward signal where most actions are penalized, ultimately discouraging exploration.
To mitigate this, we propose two complementary reward components that together provide informative and balanced feedback to guide the agent’s learning, ultimately shaping the cells in the goal area, which we define as the set of grid cells where the target height map differs from the initial (flat) configuration.

\subsubsection*{\textbf{Granular Media Shaping}}
The first component directly incentivizes reducing the discrepancy between the current and the goal state of the granular media. We consider two formulations for this reward.

The \emph{delta} reward provides feedback proportional to the reduction in distance between the current and goal configuration.
Reducing this distance leads to positive rewards.
The reward is given by:
\begin{equation}
r_{delta} =
\alpha_\textrm{c} \cdot (\hat{d}_{t \text{-} 1} - \hat{d}_{t}),
\end{equation}
where \(\alpha_\textrm{c}\) is a scaling factor and \(\hat{d}\) is the mean absolute difference between the goal height map and the truncated current height map.
It is defined as:
\begin{equation}
\hat{d} = \frac{1}{N_\text{cell}} \sum^{N_\text{cell}}_{i=1} | h_{\text{g},i} - \min(h_0, h_{\text{c},i}) |,
\end{equation}
where \(h_{\text{g},i}\) and \(h_{\text{c},i}\) denote the height at a grid cell \(i\) in \(H_\text{g}\) and \(H_\text{c}\), respectively, \(h_0\) is the initial height level of the granular medium, and \(N_\text{cell}\) is the number of grid cells.
Note that for \(\hat{d}\), the height values of the current height map \(H_\text{c}\) are cut off at the initial flat height of the granular media as the goal height map \(H_\text{g}\) only contains negative shape imprints.
In other words, piled-up granular medium is disregarded.

While \(r_{delta}\) is intuitive, its formulation can discourage exploration by penalizing intermediate actions that happen to increase the distance to the goal, but lead to it on the long run.
To address this challenge, we draw inspiration from reward shaping techniques that have been employed for 6D object reposing~\cite{petrenkoDexPBTScalingDexterous2023a}.
The so-called \emph{progressive} reward, rewards the agent for making progress relative to the best configuration reached so far within the current episode, rather than relative to the immediately preceding time step.
In this case, the agent receives a reward when it changes the granular medium into a configuration that is closer to the target configuration than all other ones that have been reached before within the episode. 
Likewise, we penalize the agent for reaching configurations further from the target than the furthest one reached so far within the same episode.
The progressive reward is defined as:
\begin{equation}
r_{prog} =
\alpha_\textrm{c} \cdot \max (\hat{d}_\textrm{closest} - \hat{d}, 0)
- \alpha_\textrm{f} \cdot \min(\hat{d}^o -\hat{d}^o_\textrm{furthest}, 0),
\end{equation}
where \(\hat{d^o}\) is the mean absolute difference of the current and the goal height map outside the goal area, \(\cdot_\textrm{closest}\) and \(\cdot_\textrm{furthest}\) denote the closest and furthest reached distances inside the current episode, respectively.
This formulation avoids cycles of positive return.

For both proposed rewards \(r_{delta}\) and \(r_{prog}\), we use the scaling factors \(\alpha_\textrm{c} = 5,000\) and \(\alpha_\textrm{f} = 1,000\).

\subsubsection*{\textbf{Goal Area Movement}}
To encourage the agent to move its EE towards the region of interest and to stay within this area,  we add a further reward term using the distance of the EE to the goal area.
The agent is incentivized to bring its EE close to this area using a distance-based penalty, with a binary bonus for reaching the region.
Formally, we define the reward as:
\begin{equation}
r_\text{m} = -\tanh(\alpha_\textrm{m} \cdot d_\textrm{m}) + \mathds{1}_\textrm{reached},
\end{equation}
where \(d_\textrm{m}\) is the minimum Euclidean distance between the EE and the goal area, computed as visualized in \figref{fig:reward_shaping} and \(\mathds{1}_\textrm{reached}\) is an indicator function that returns 1 when the EE is inside the goal area and 0 otherwise. We use a value of \(\alpha_\textrm{m} = 10\) to control the steepness of the distance-based penalty.

The total reward is computed as the sum of the goal movement reward and the shaping reward:
\begin{equation}
\label{eq:reward}
r = r_\text{m} + r_\text{s},
\end{equation}
where \(r_\text{s} \ \{ r_{delta}, r_{prog} \}\).

%%%%%%%%%%%%%%%%%%%%%%%%%%%%%%%%%%%%%%%%%%%%%%%%%%%%%%%%%%%%%%%%%%%%%%%%%%%%%%%%
\section{Experimental Evaluation}
\label{sec:exp}
To demonstrate the performance of our RL approach to shape granular media compared to two baselines, we performed experiments with different goal height maps.
We further tested different reward formulations and environment state observability modes.
Additionally, we conducted an ablation study of the feature extractor and the choice of RL algorithm.
More details of our method, the supplementary video, and our code are available on the paper~website\footnote{Paper website: \href{https://humanoidsbonn.github.io/granular_rl/}{https://humanoidsbonn.github.io/granular\_rl/}}.

%%%%%%%%%%%%%%%%%%%%%%%%
\subsection{Baselines}
\label{subsec:baselines}
We implemented two different baseline approaches for comparison, which we detail in this section. 

\subsubsection*{\bfseries{Random Baseline}}
\label{subsubsec:random_baseline}
For the random baseline \textbf{(RAND)}, at the start of each evaluation episode we place the robot's EE at a random start position uniformly sampled within the goal mask \(M_\mathrm{g}\) and at the surface of the granular medium.
The robot then executes random actions for \(N_{\mathrm{ep}}\) time steps, after which the episode strictly terminates.

\subsubsection*{\bfseries{Boustrophedon Coverage Path Planning Baseline}}
\label{subsubsec:cpp_baseline}
For this baseline \textbf{(B-CPP)}, we utilize Boustrophedon decomposition~\cite{chosetCoverageKnownSpaces2000} combined with Coverage Path Planning~(CPP), following the approach of Terenzi and Hutter~\cite{terenziAutonomousExcavationPlanning2024}.  
First, we perform a flood fill on the Boolean goal mask \(M_\text{g}\) to extract connected regions.
We then compute the centroid of each region and solve the traveling salesman problem on these centroids by performing a greedy nearest-neighbor search~\cite{rosenkrantzApproximateAlgorithmsTraveling1974}.
Within each region, we generate parallel sweep lines spaced by the EE’s footprint and alternate direction on each pass to minimize the traversal overhead.
Note that if the footprint partially covers a cell, it is rounded to a whole cell.
For each grid coordinate in a sweep, we extract the target height and assemble a sequential list of waypoints.  
During an evaluation episode, the robot moves its EE to each waypoint following the order of the list, terminating once the full plan is executed.
Hence, the length of an episode is defined by the number of waypoints.

%%%%%%%%%%%%%%%%%%%%%%%%
\subsection{Experimental Setup}
\label{subsec:setup}
We evaluate the effectiveness of our policies, by randomly selecting out of 400 distinct goal height maps in the beginning of each episode.
We distinguish between heuristic baselines, policies trained in a privileged setting with full state observability, and policies using observations that are obtainable in the real world, consequently suffering, e.g., from occlusions.
\tabref{tab:rl_training} lists all relevant RL training parameters.

For all experiments, we use the 6-DoF robotic arm UR5e, equipped with a custom, cubic end-effector (with dimensions of \mbox{\(2\)\,\texttimes\,\(2\)\,\texttimes\,\(\SI{15}{\cubic\centi\meter}\)),} and an external ZED~2i camera, \mbox{as shown in \figref{teaser}}.
However, our approach is applicable to any robotic arm, since our policies learn EE movement increments instead of robot joint positions.
Furthermore, we rely on the RL algorithm implementations of Stable-Baselines3~\cite{raffinStableBaselines3ReliableReinforcement2021},
and the OpenAI Gym toolkit~\cite{brockmanOpenAIGym2016a}, as well as robosuite~\cite{zhuRobosuiteModularSimulation2025} that is based on MuJoCo~\cite{todorovMuJoCoPhysicsEngine2012} for simulation, and the Robotics Toolbox for Python~\cite{corkeNotYourGrandmothers2021}.
\begin{table}[b]
	\centering
	\caption{RL notations and training settings.}
        \label{tab:rl_training}
	\begin{tabularx}{\linewidth}{llX}
		\toprule[\lightrulewidth]
		Notation & TQC\,/\,SAC\,/\,TD3 & Description \\
		\midrule[\lightrulewidth] 
		\(N_\text{ep}\) & 40 & Number of steps per episode\\
		\(N_\text{crit}\) & 2 & Number of critics\\
		\(B_\textrm{E}\) & \(1 \cdot 10^{5}\) & Experience replay buffer size\\
		\(B_\textrm{G}\) & \num{256} &  Minibatch for each gradient update\\
		\(l_\textrm{T}\) & \(3 \cdot 10^{-4}\)  & Learning rate\\
		\(\gamma\) & 0.99 & Discount factor\\
		\(\sigma_{\epsilon_\pi}\) & 0.2 & Std. deviation of exploration noise \(\epsilon_\pi\) \\
		\(\tau\) & 0.005 & Soft update coefficient\\		
		\bottomrule[\lightrulewidth]
	\end{tabularx}
\end{table}
%
% \newlength{\goalSubfigWidth}
% \setlength{\goalSubfigWidth}{0.155\textwidth}
\newcommand{\goalSubfigWidth}{0.14\textwidth}
\newcommand{\goalTopRowScale}{0.95\textwidth}
\newcommand{\goalSubfigSpace}{0.1cm}

\newcommand{\goalImageWidth}{2.0cm}
\newcommand{\frescpImageWidth}{2.2cm}

\begin{figure}[t]
    \centering
        \begin{minipage}{\linewidth}
            \begin{minipage}{\goalImageWidth}
                \centering
                \small{Achieved~3D}
                \small{render~of~\(H_\text{c}\)}
                %\vspace{-1cm}
            \end{minipage}
            \hfill
            \begin{minipage}{\goalImageWidth}
                \centering
                \small{Achieved}
                \small{height~map~\(H_\text{c}\)}
                % \vspace{-1cm}
            \end{minipage}
            \hfill
            \begin{minipage}{\goalImageWidth}
                \centering
                \small{Goal}
                \small{height~map~\(H_\text{g}\)}
                % \vspace{-1cm}
            \end{minipage}
                \hfill
            \hspace{\frescpImageWidth}
        \end{minipage}
        
        \vspace{0.1cm}
        
        \begin{minipage}{\linewidth}
            \includegraphics[width=\goalImageWidth]      {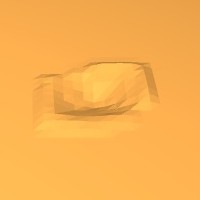}
            \hfill
            \includegraphics[width=\goalImageWidth,angle=180,origin=c]{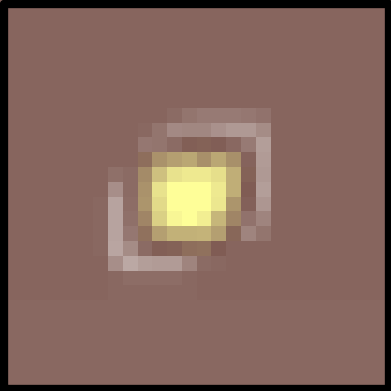}
            \hfill
            \includegraphics[width=\goalImageWidth,angle=180,origin=c]{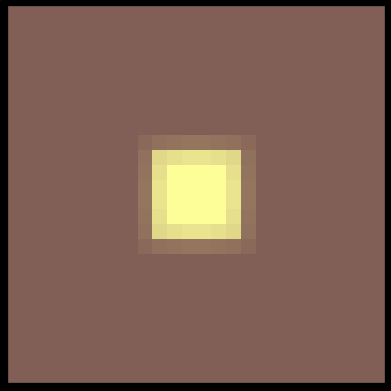}
            \hfill
            \hspace{\frescpImageWidth}
        \end{minipage}

        \vspace{0.1cm}
        
        \begin{minipage}{\linewidth}
            \includegraphics[width=\goalImageWidth]      {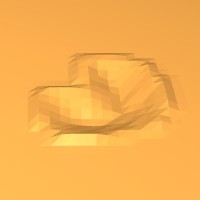}
            \hfill
            \includegraphics[width=\goalImageWidth,angle=180,origin=c]{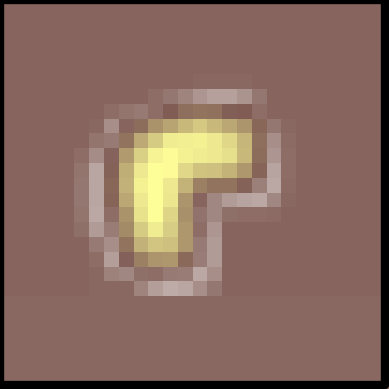}
            \hfill
            \includegraphics[width=\goalImageWidth,angle=180,origin=c]{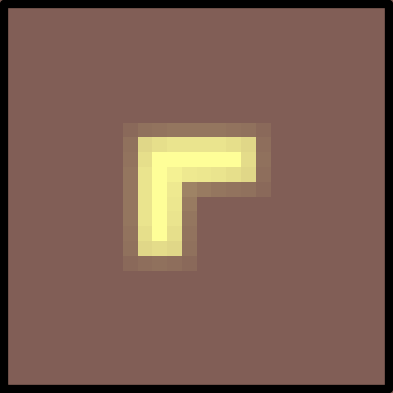}
            \hfill
            \hspace{\frescpImageWidth}
        \end{minipage}

        \vspace{0.1cm}
        
        \begin{minipage}{\linewidth}
            \includegraphics[width=\goalImageWidth]      {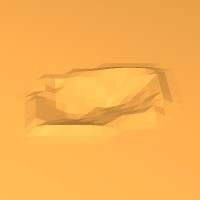}
            \hfill
            \includegraphics[width=\goalImageWidth,angle=180,origin=c]{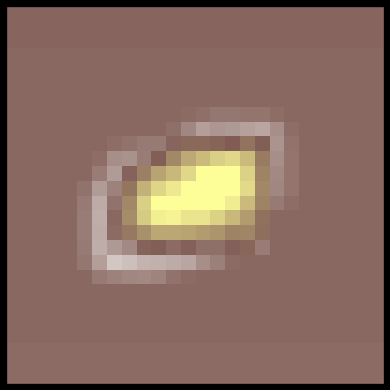}
            \hfill
            \includegraphics[width=\goalImageWidth,angle=180,origin=c]{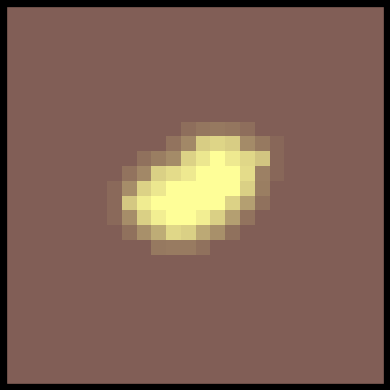}
            \hfill
            \begin{minipage}{\frescpImageWidth}
                \centering
                \small{Fresco~fragment}
                \vspace{-1cm}
            \end{minipage}
        \end{minipage}
        
        \vspace{0.1cm}
        
        \begin{minipage}{\linewidth}
            \includegraphics[width=\goalImageWidth]      {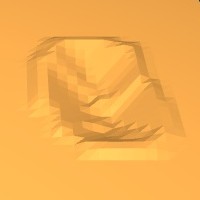}
            \hfill
            \includegraphics[width=\goalImageWidth,angle=180,origin=c]{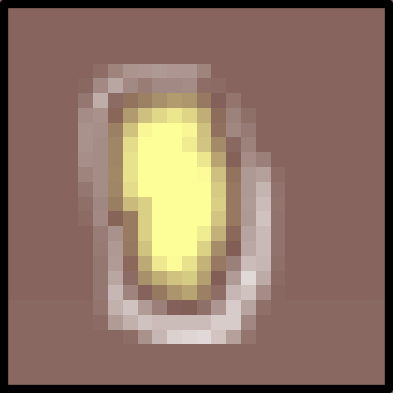}
            \hfill
           \includegraphics[width=\goalImageWidth,angle=180,origin=c]{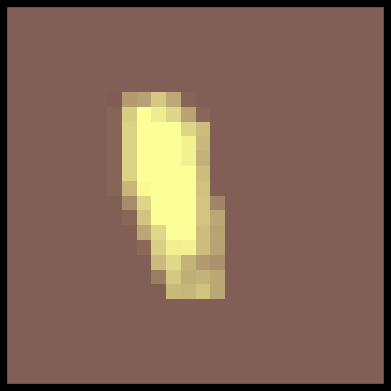}
            \hfill
            \includegraphics[width=\frescpImageWidth]{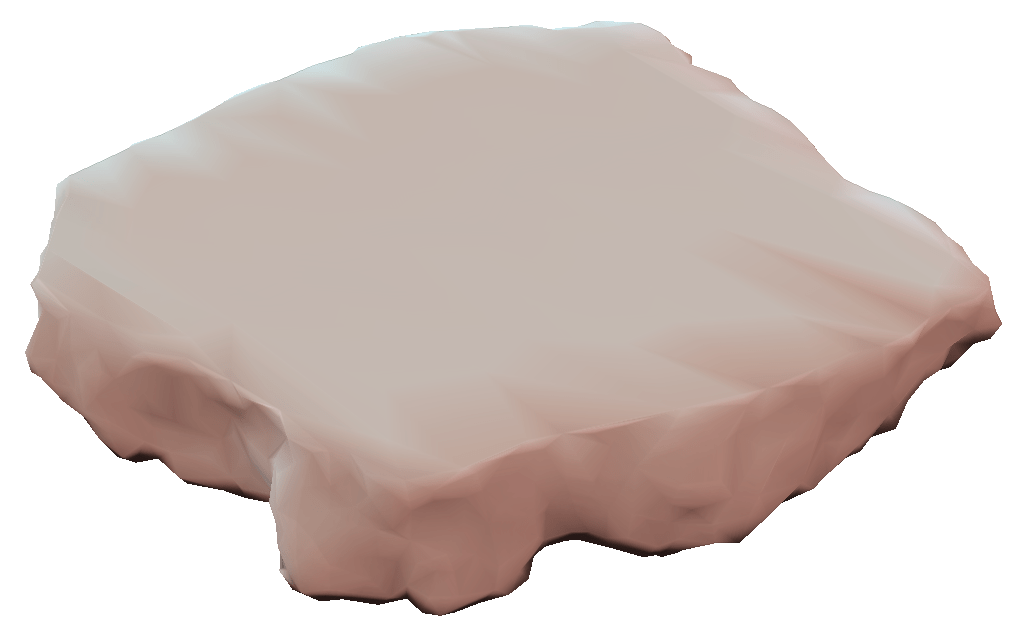}
        \end{minipage}
    \caption{From left to right, we show the reconstructed 3D scene in simulation, the reconstructed height map after the manipulation, and the goal height map the agent aims to achieve. From top to bottom, the given goal shapes are an exemplary rectangle, an L, a polygon, and a fresco fragment's negative.}
    \label{fig:goal_examples}
\end{figure}

\newcommand{\rolloutSubfigWidth}{0.14\textwidth}
\newcommand{\rolloutTopRowScale}{0.95\textwidth}
\newcommand{\rolloutSubfigSpace}{0.1cm}

\newcommand{\rolloutImageWidth}{2.18cm}

\begin{figure*}[t]
    \centering
    \begin{subfigure}{0.8\textwidth}
        \begin{minipage}{0.8\textwidth}
            \includegraphics[trim=9cm 2cm 6cm 0cm, clip, width=\rolloutImageWidth]{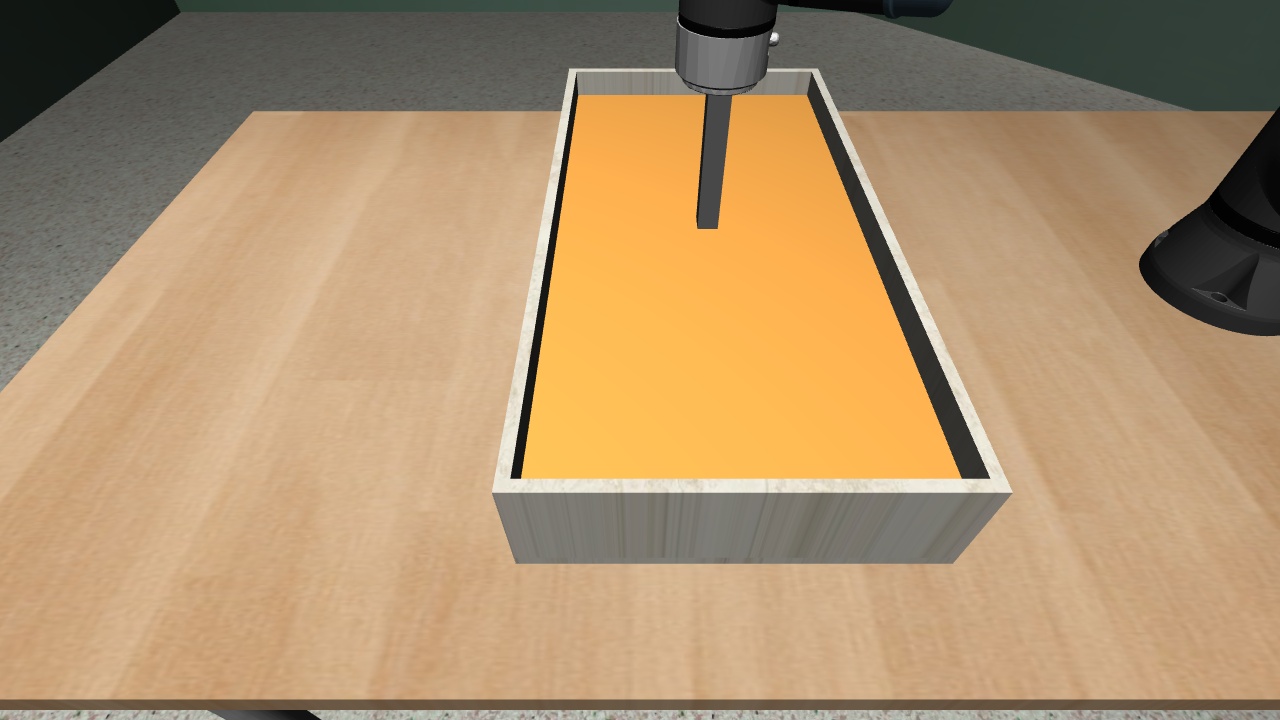}
            \hfill
            \includegraphics[trim=9cm 2cm 6cm 0cm, clip, width=\rolloutImageWidth]      {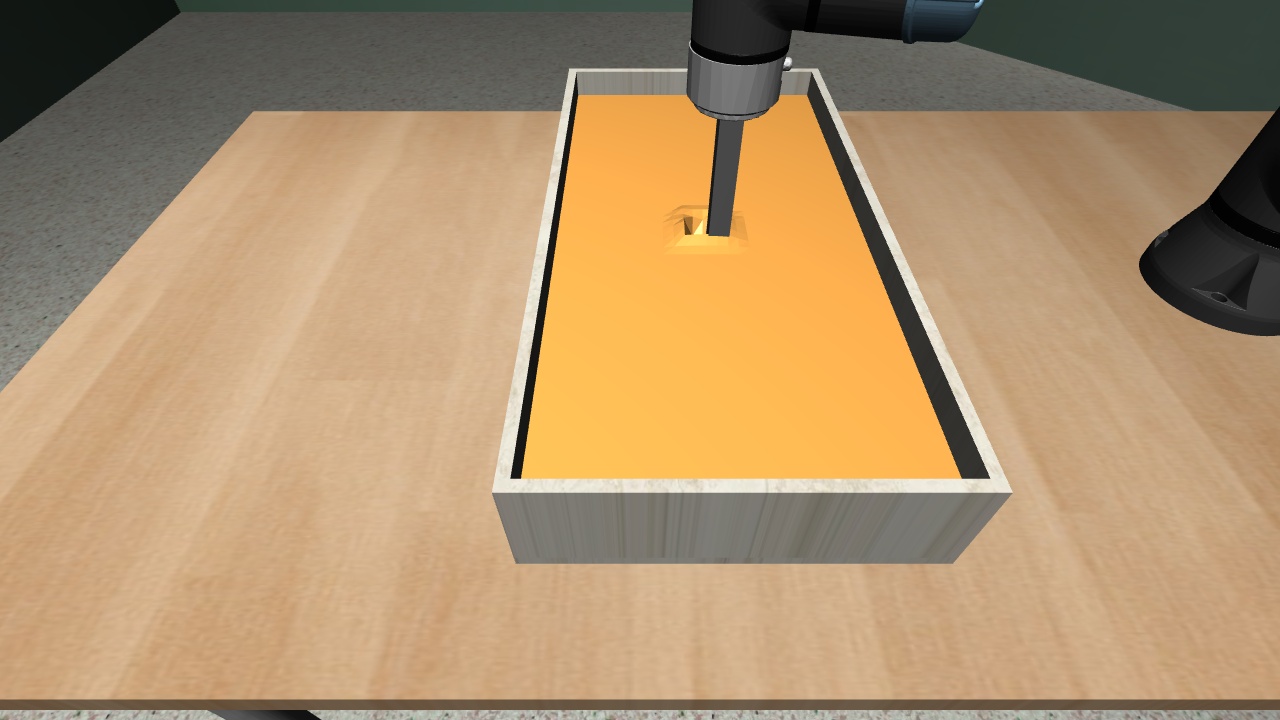}
            \hfill
            \includegraphics[trim=9cm 2cm 6cm 0cm, clip, width=\rolloutImageWidth]{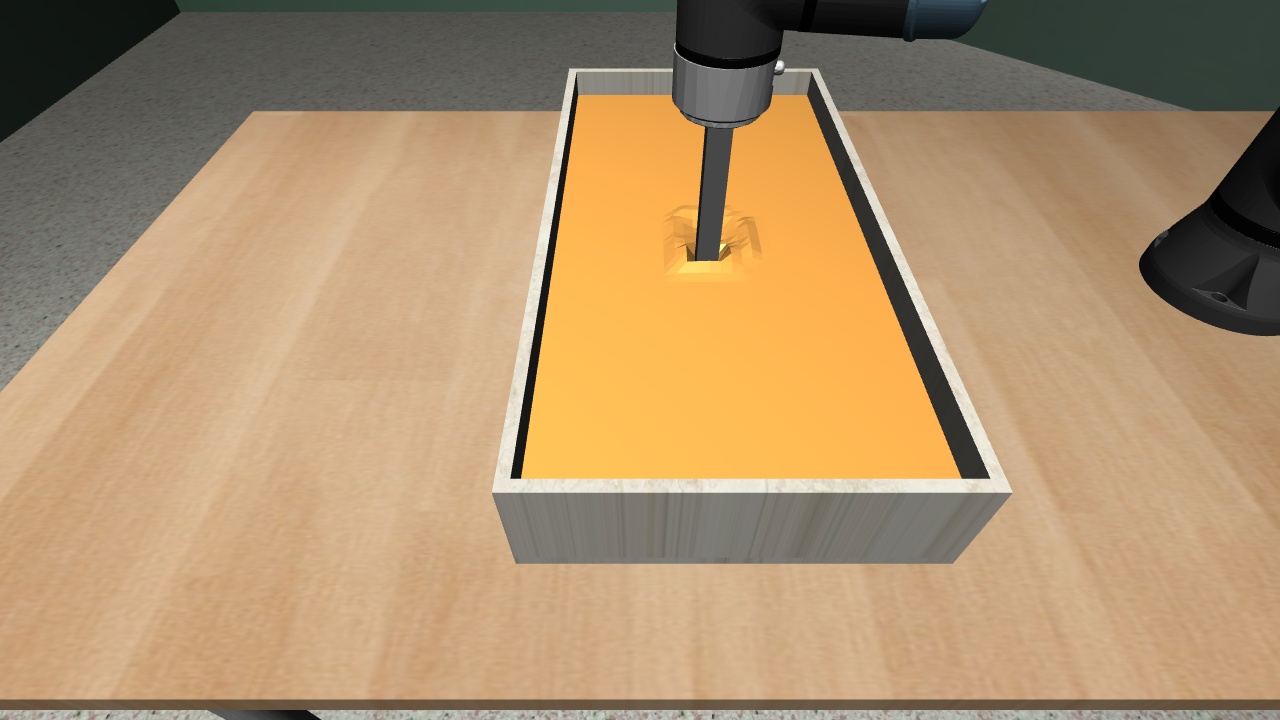}
            \hfill
            \includegraphics[trim=9cm 2cm 6cm 0cm, clip, width=\rolloutImageWidth]{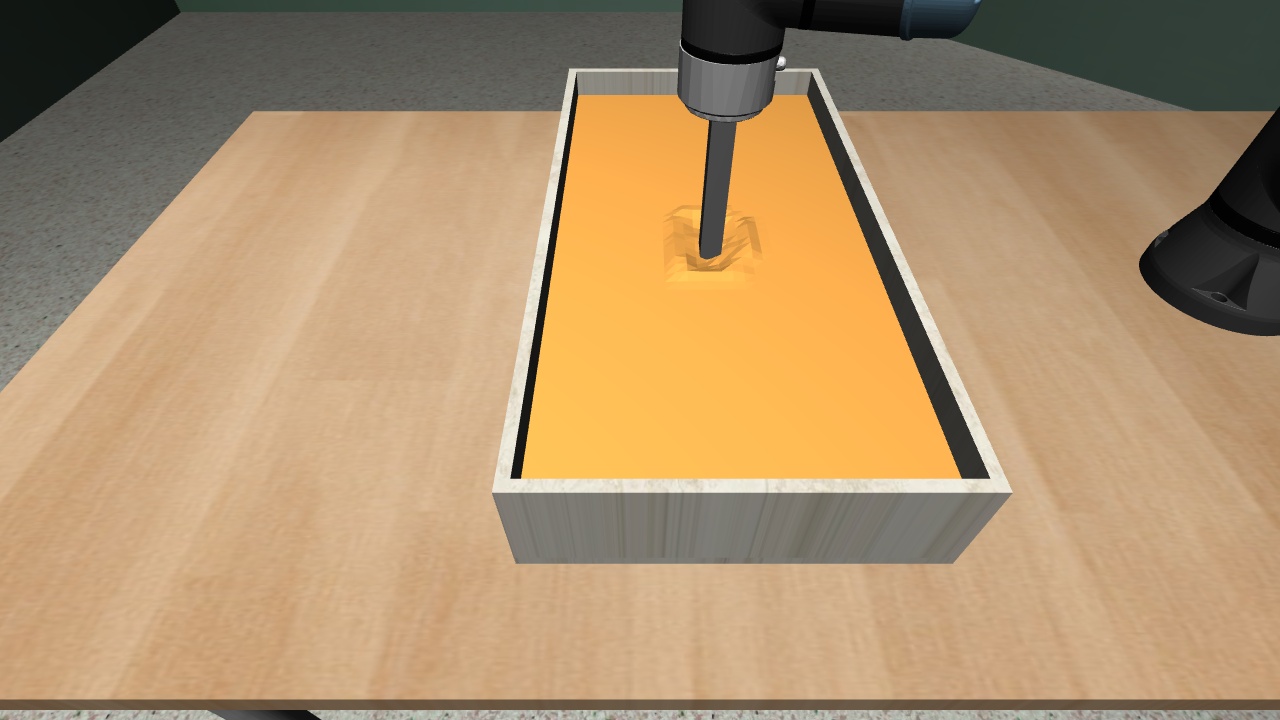}
            \hfill
            \includegraphics[trim=9cm 2cm 6cm 0cm, clip, width=\rolloutImageWidth]{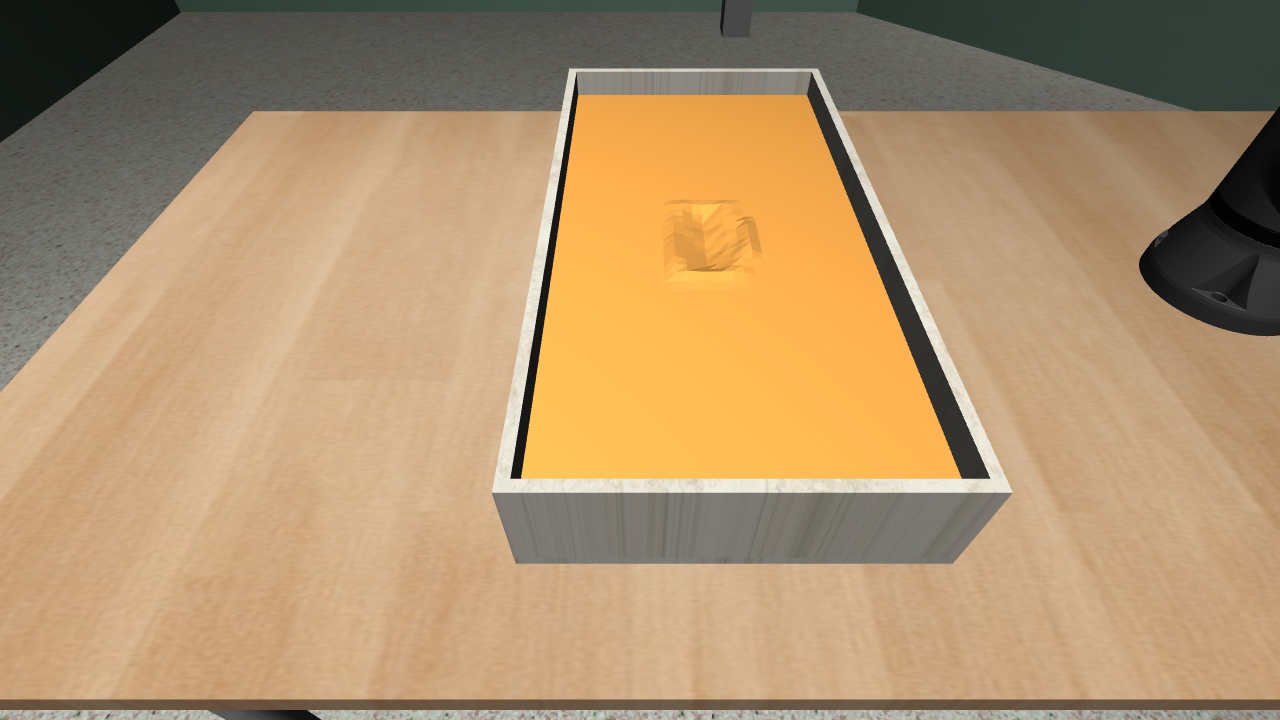}
        \end{minipage}
        \hfill
        \begin{minipage}{\rolloutImageWidth}
            \centering
            \small{Goal}
            
            \small{height~map~\(H_\text{g}\)}
            \vspace{-1cm}
        \end{minipage}
        
        \vspace{0.1cm}
        
        \begin{minipage}{0.8\textwidth}
            \includegraphics[width=\rolloutImageWidth,angle=180,origin=c]{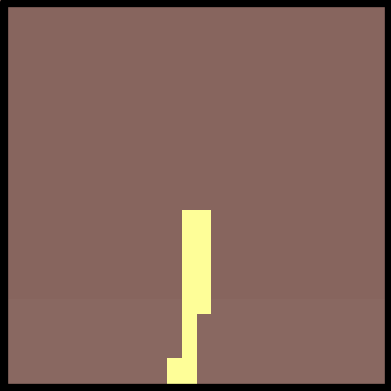}
            \hfill
            \includegraphics[width=\rolloutImageWidth,angle=180,origin=c]{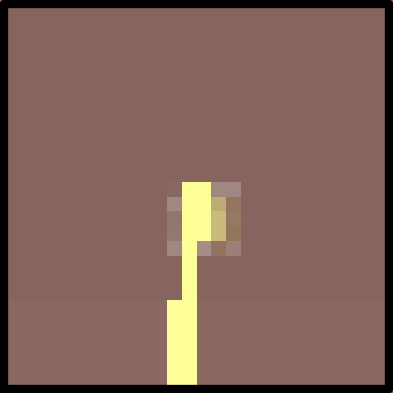}
            \hfill
            \includegraphics[width=\rolloutImageWidth,angle=180,origin=c]{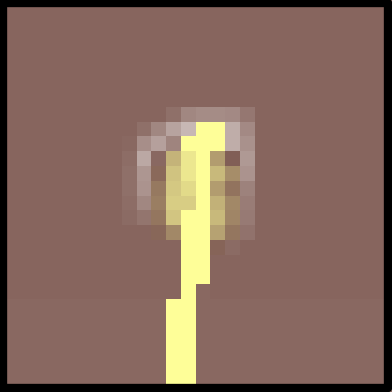}
            \hfill
            \includegraphics[width=\rolloutImageWidth,angle=180,origin=c]{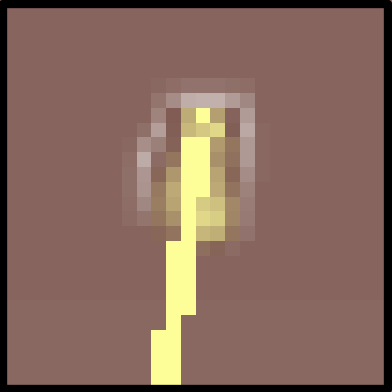}
            \hfill
            \includegraphics[width=\rolloutImageWidth,angle=180,origin=c]{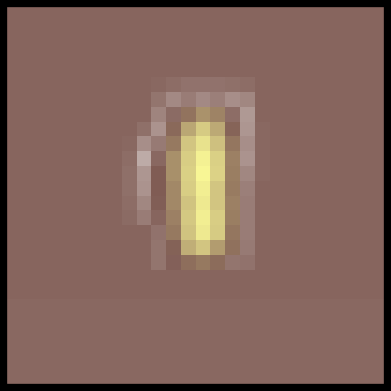}
        \end{minipage}
        \hfill
        \begin{minipage}{\rolloutImageWidth}
            \centering
            \includegraphics[width=\textwidth,angle=180,origin=c]{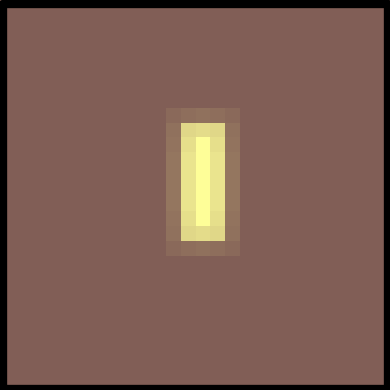}
        \end{minipage}
        \subcaption{The agent is forming the given goal height map in the \textbf{simulated environment} (top). The resulting current reconstructed height maps (bottom) include occlusions of the end-effector. At the start of the episode (left) the granular medium is perfectly flat and in the end the desired goal shape is visible in it (right).}
        \label{fig:simulation}
    \end{subfigure}

    \vspace{0.1cm}

    \begin{subfigure}{0.8\textwidth}
        \begin{minipage}{0.8\textwidth}
            \includegraphics[trim=12cm 3cm 10cm 0, clip, width=\rolloutImageWidth]{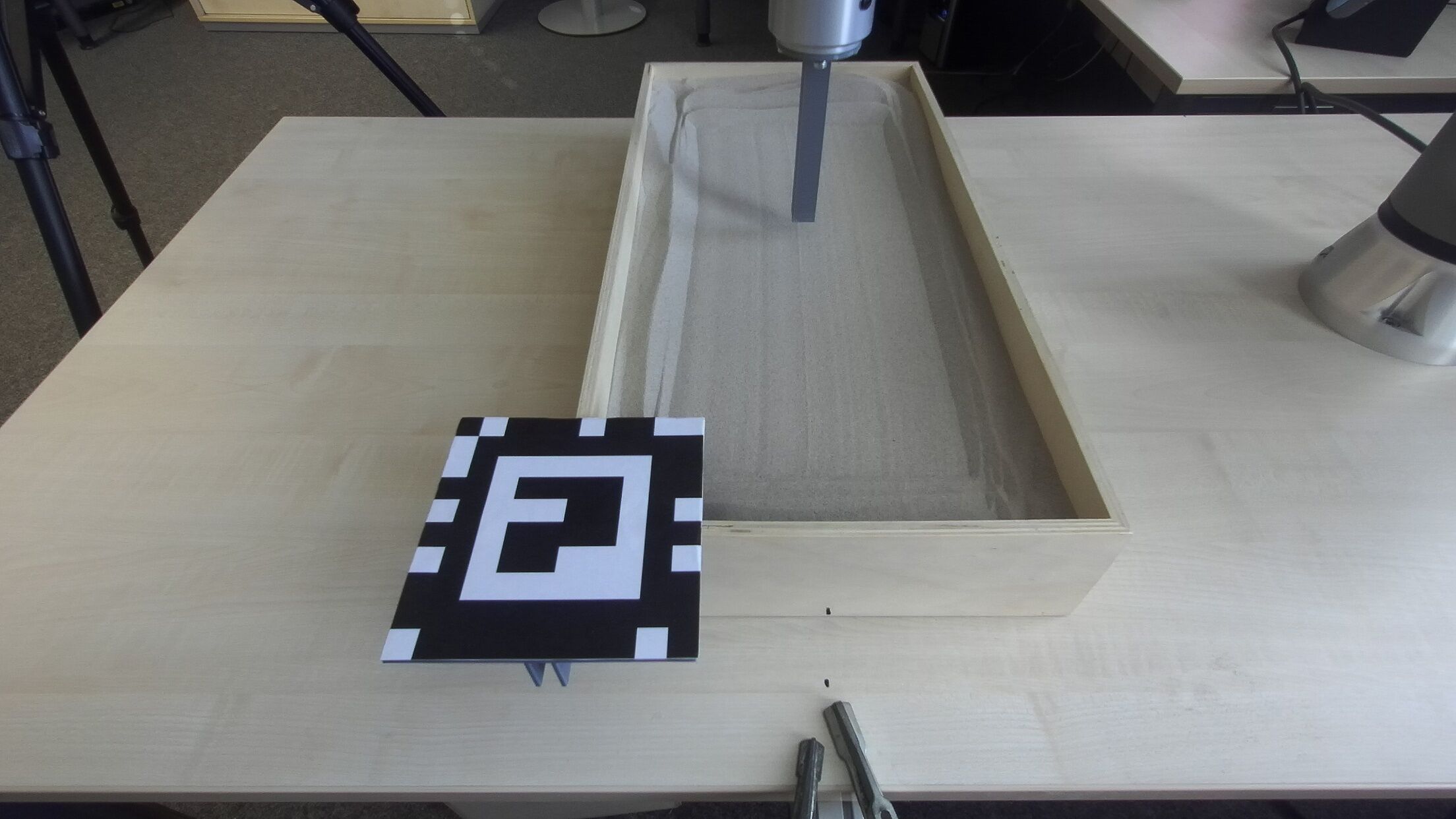}
            \hfill
            \includegraphics[trim=12cm 3cm 10cm 0, clip, width=\rolloutImageWidth]{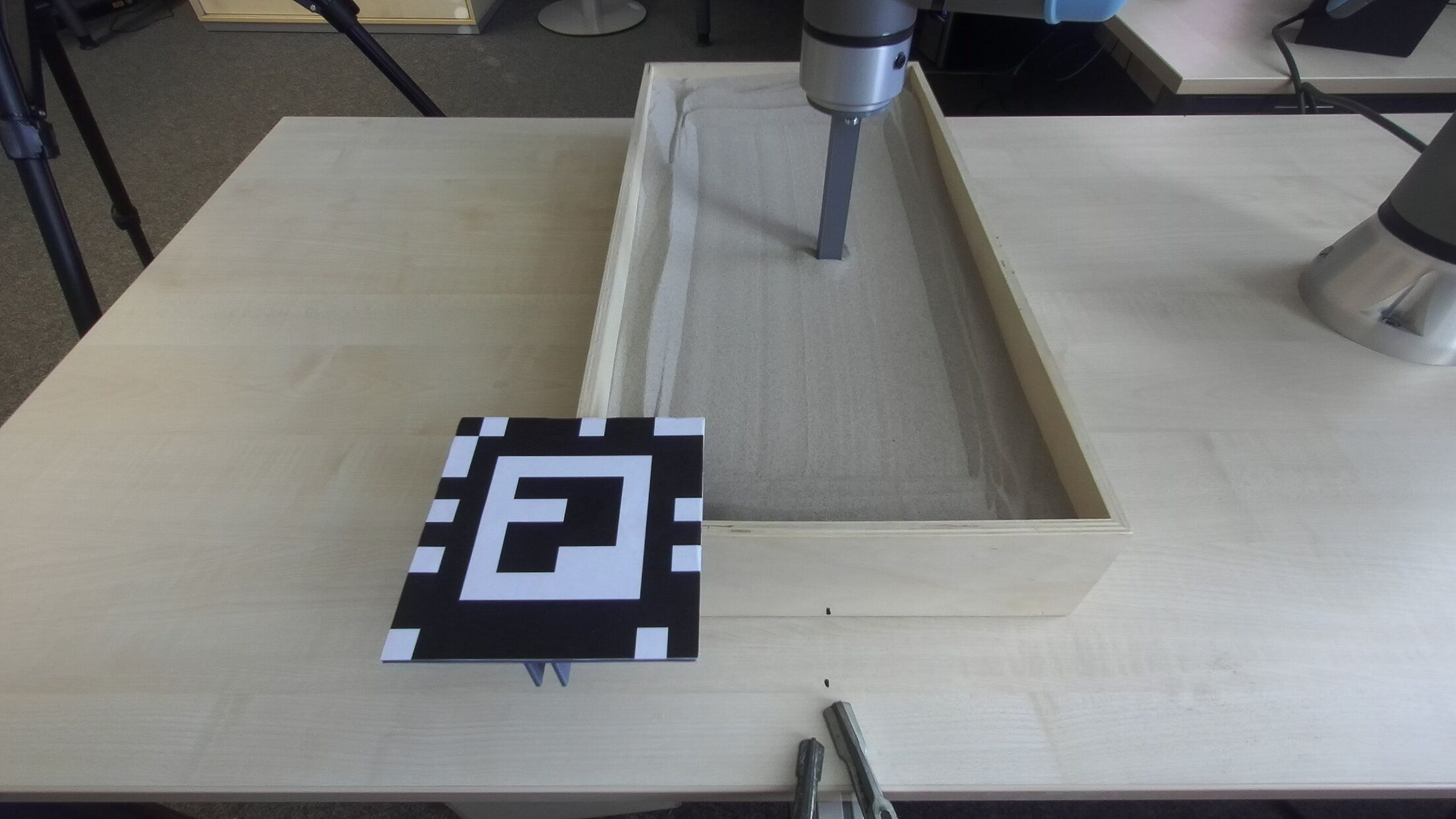}
            \hfill
            \includegraphics[trim=12cm 3cm 10cm 0, clip, width=\rolloutImageWidth]{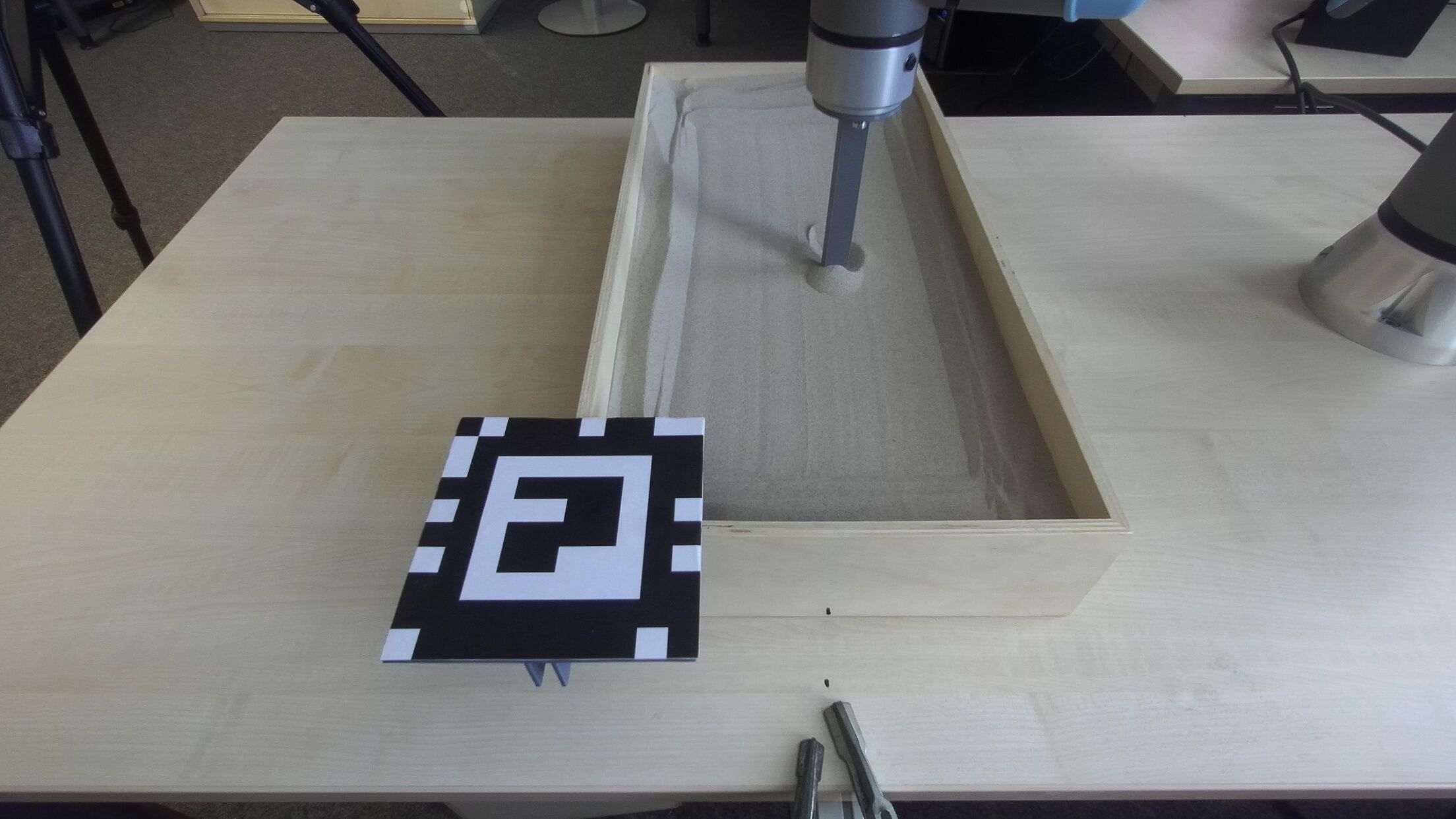}
            \hfill
            \includegraphics[trim=12cm 3cm 10cm 0, clip, width=\rolloutImageWidth]{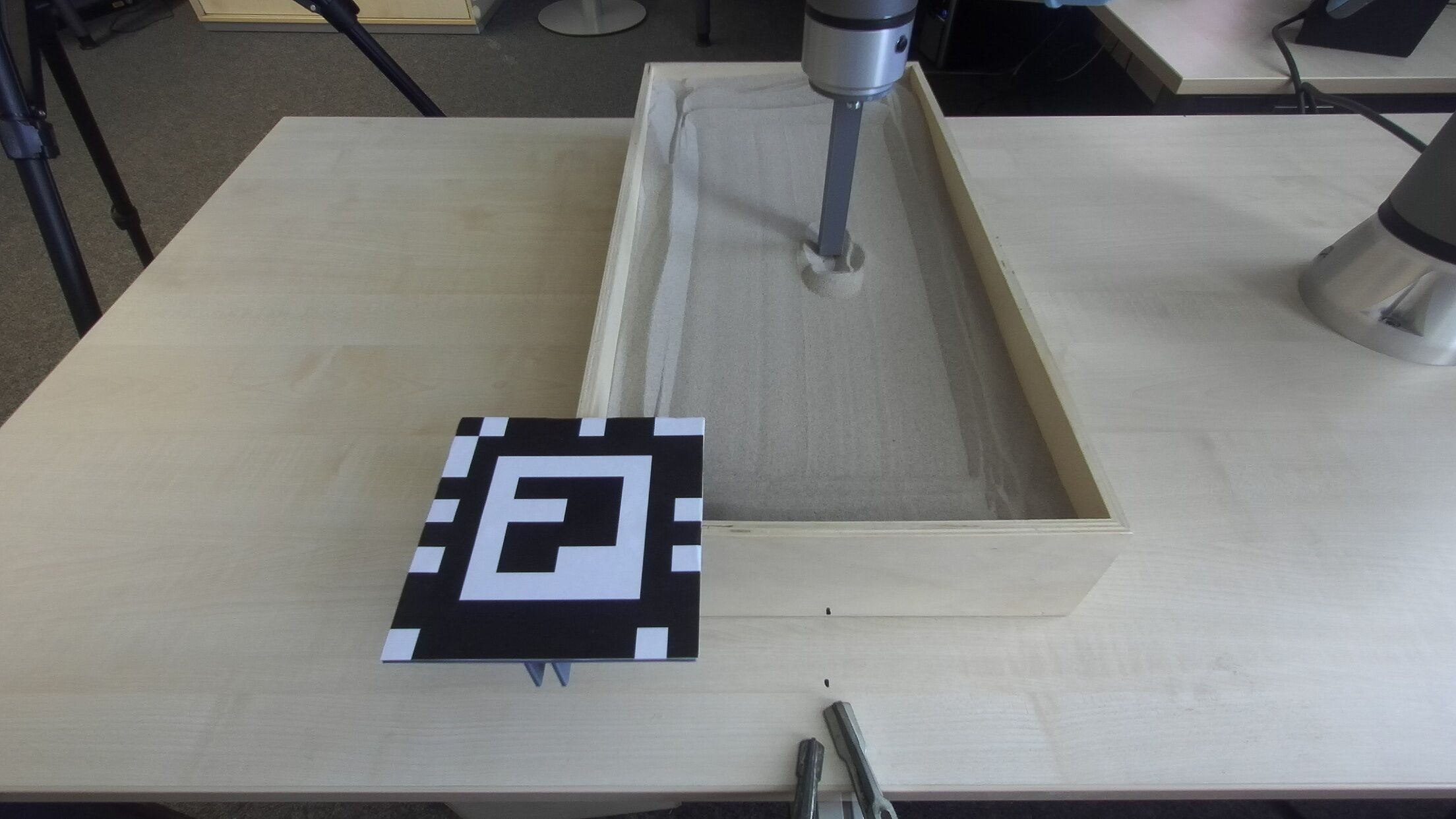}
            \hfill
            \includegraphics[trim=12cm 3cm 10cm 0, clip, width=\rolloutImageWidth]{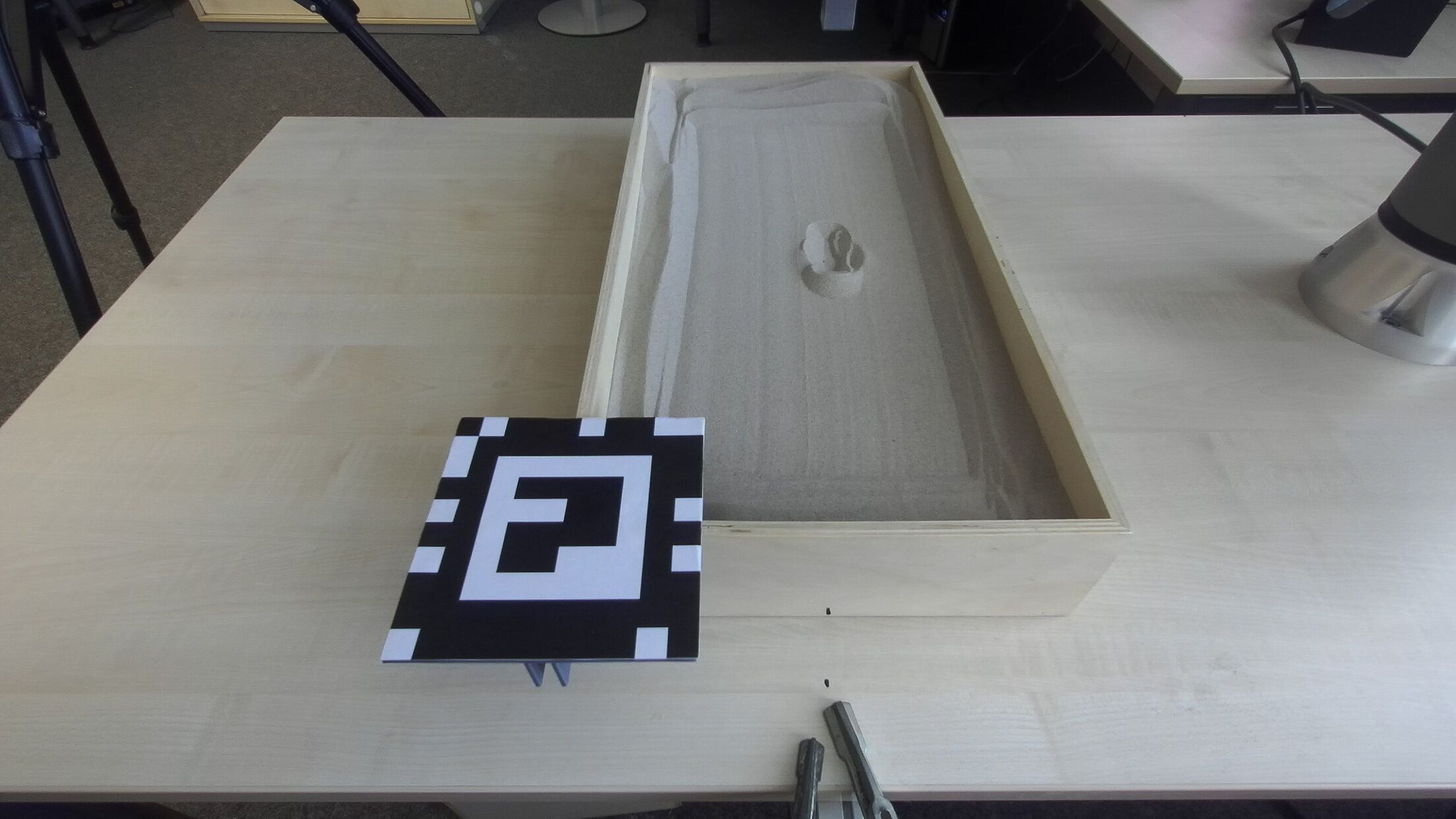}
        \end{minipage}
        \hfill
        \begin{minipage}{\rolloutImageWidth}
            \centering
            %\textsf{\footnotesize{None}}
            %\vspace{-1cm}
        \end{minipage}
        
        \vspace{0.1cm}
        
        \begin{minipage}{0.8\textwidth}
            \includegraphics[trim=9cm 2cm 6cm 0cm, clip, width=\rolloutImageWidth]{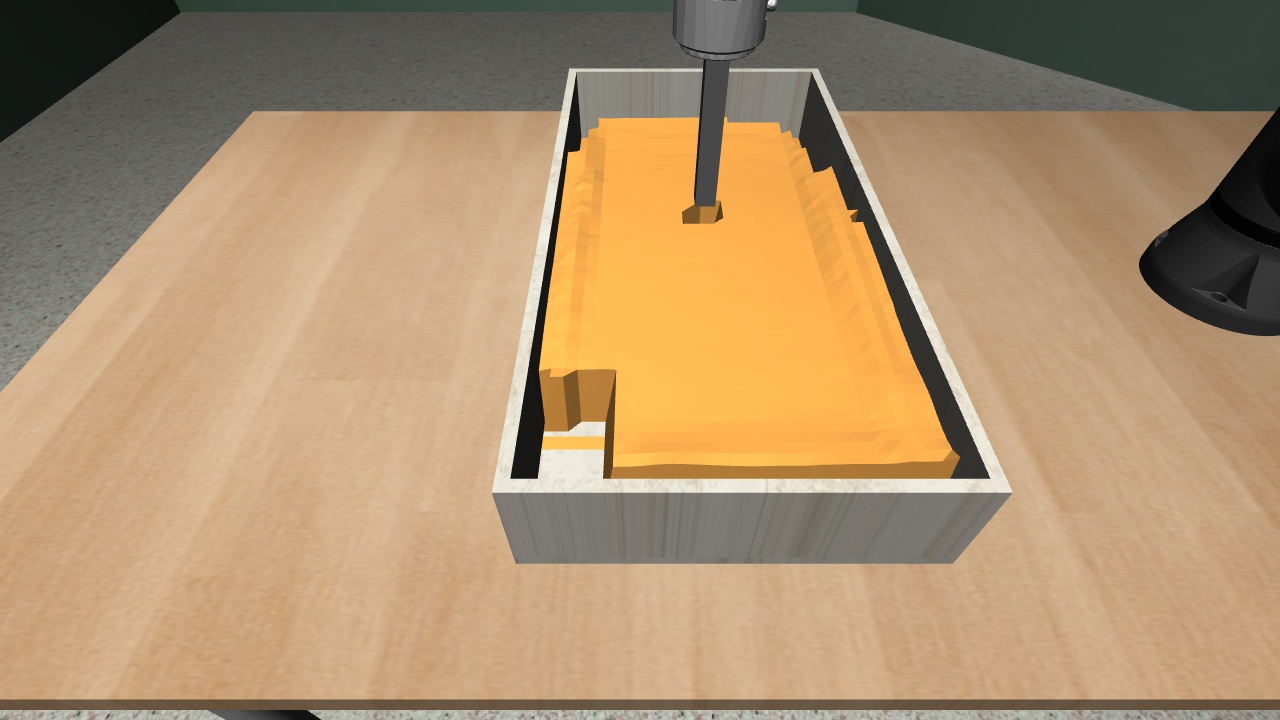}
            \hfill
            \includegraphics[trim=9cm 2cm 6cm 0cm, clip, width=\rolloutImageWidth]      {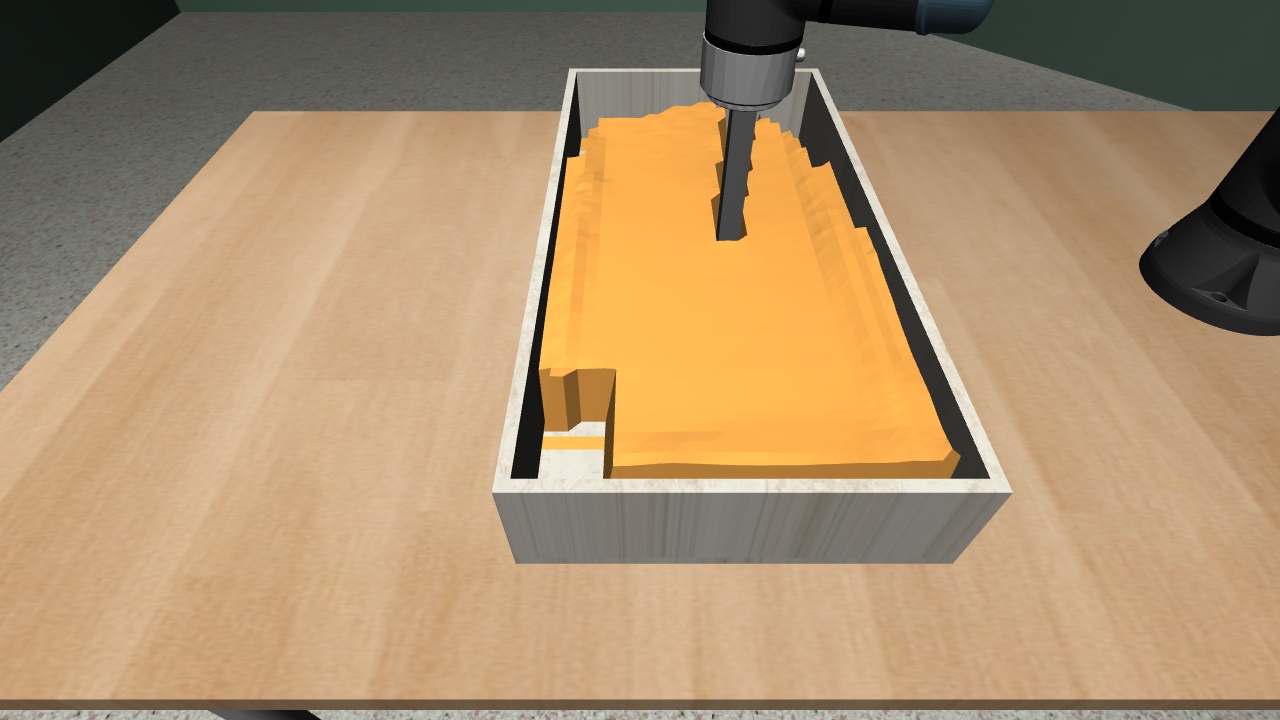}
            \hfill
            \includegraphics[trim=9cm 2cm 6cm 0cm, clip, width=\rolloutImageWidth]{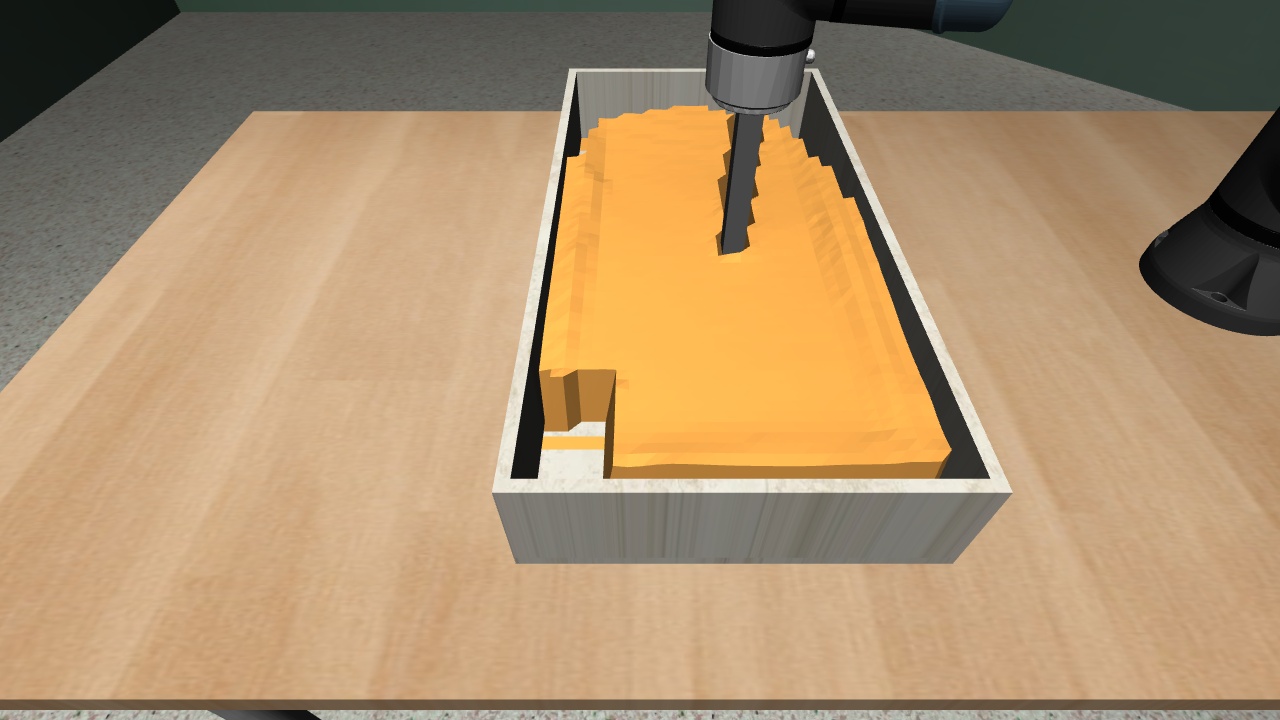}
            \hfill
            \includegraphics[trim=9cm 2cm 6cm 0cm, clip, width=\rolloutImageWidth]{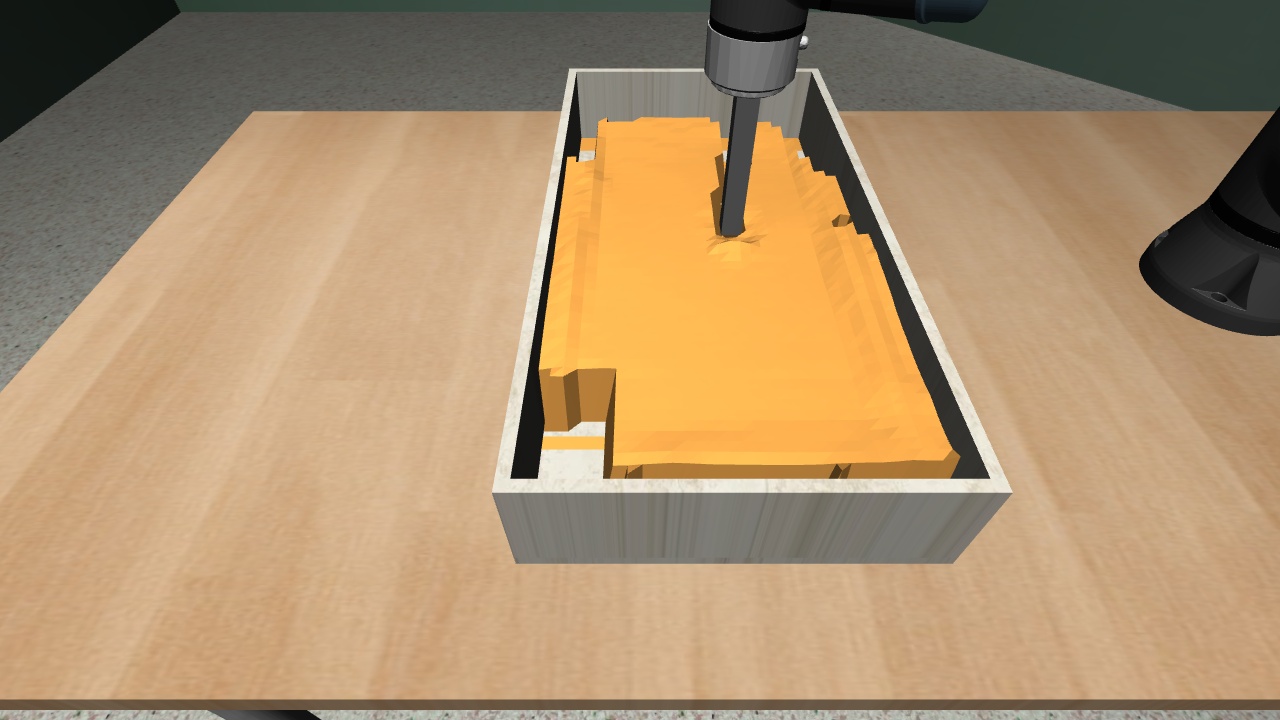}
            \hfill
            \includegraphics[trim=9cm 2cm 6cm 0cm, clip, width=\rolloutImageWidth]{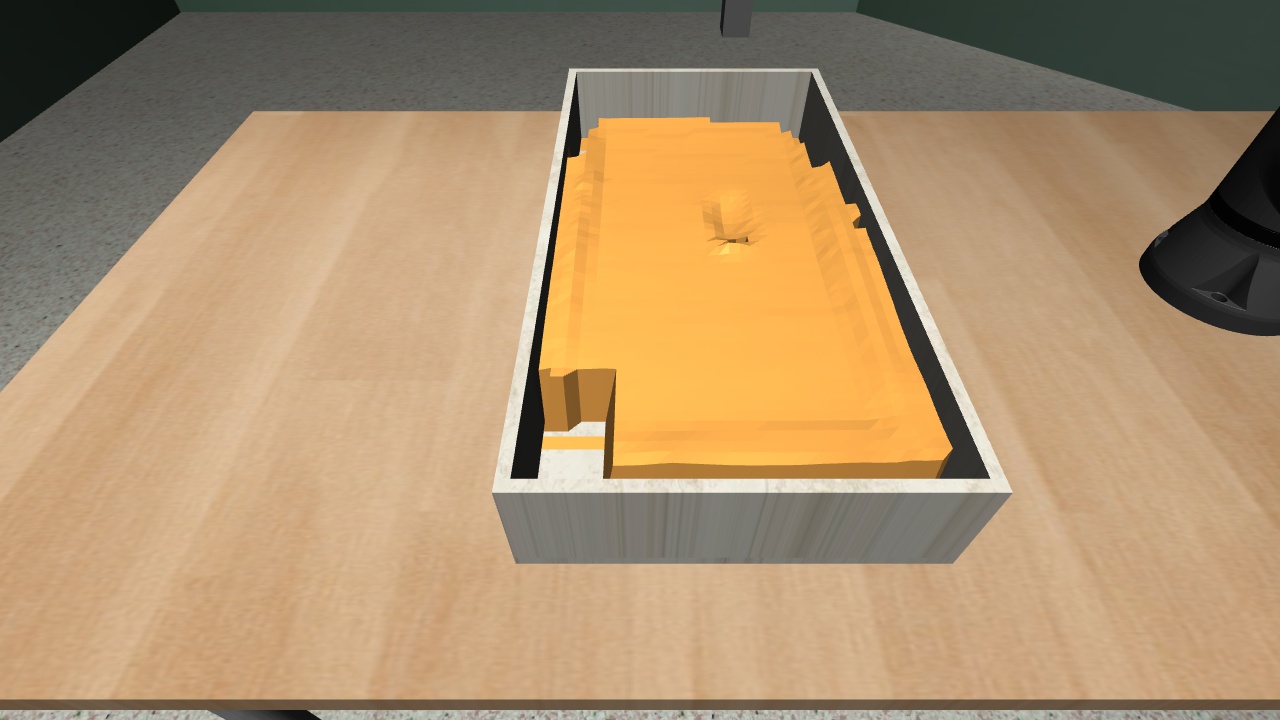}
        \end{minipage}
        \hfill
        \begin{minipage}{\rolloutImageWidth}
            \centering
            \small{Goal}
            
            \small{height~map~\(H_\text{g}\)}
            \vspace{-1cm}
        \end{minipage}
        
        \vspace{0.1cm}
        
        \begin{minipage}{0.8\textwidth}
            \includegraphics[width=\rolloutImageWidth,angle=180,origin=c]{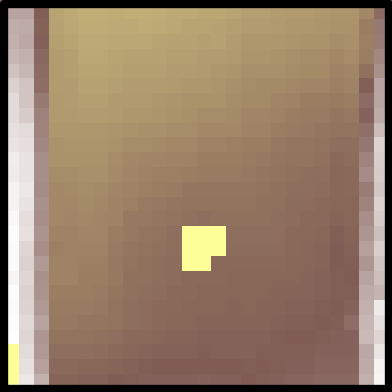}
            \hfill
            \includegraphics[width=\rolloutImageWidth,angle=180,origin=c]{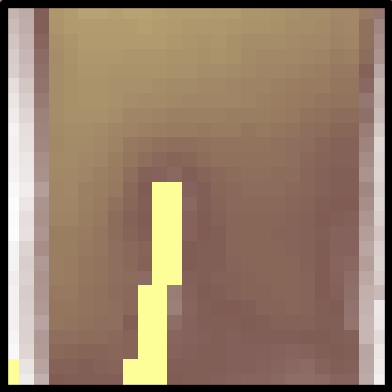}
            \hfill
            \includegraphics[width=\rolloutImageWidth,angle=180,origin=c]{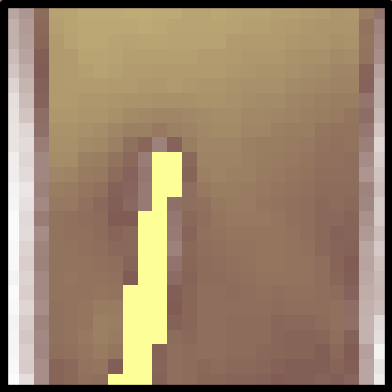}
            \hfill
            \includegraphics[width=\rolloutImageWidth,angle=180,origin=c]{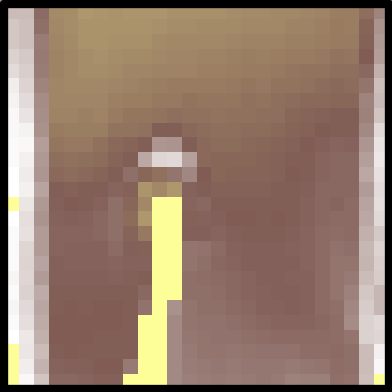}
            \hfill
            \includegraphics[width=\rolloutImageWidth,angle=180,origin=c]{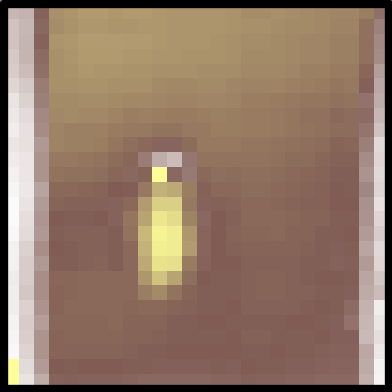}
        \end{minipage}
        \hfill
        \begin{minipage}{\rolloutImageWidth}
            \centering
            \includegraphics[width=\textwidth,angle=180,origin=c]{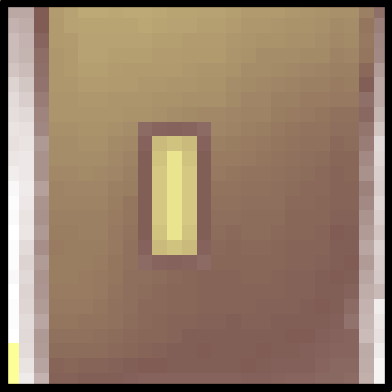}
        \end{minipage}
        \subcaption{We deploy the agent to the \textbf{real robot} in a zero-shot manner, without any additional training. From top to bottom, we show the raw camera view of the box filled with the granular medium, the reconstructed 3D scene in simulation, and the reconstructed height map that the agent observes, while it manipulates the granular medium to match the desired goal configuration. Note that the granular medium is not perfectly flat like in the simulated scenario. Despite noisy observations from the depth camera, the agent is able to create the desired rectangle shape in the real world.}
    \label{fig:transfer}
    \end{subfigure}
    
    \caption{Manipulation of the granular medium by a visual, goal-conditioned RL agent in simulation (top) and on the real robotic system (bottom).}
    \label{fig:rollouts}
\end{figure*}

\subsubsection*{\textbf{Goal Height Maps}}
We evaluate on a wide range of different shapes represented as goal height maps: 100 rectangles, 100 L-shapes, 100 polygons, and 100 negatives of archaeological fresco fragments.
The goal area of the shapes are up to \(10\)\,\texttimes\,\(\SI{10}{cm}\) with varying heights of up to \SI{3}{cm}.
We randomly place the shapes within the granular media, resulting in various configurations of each shape.
All target shapes are designed such that they are not achievable by the agent executing a single stroke through the granular media.

To generate the fresco fragment goals, we place three-dimensional scans of fresco fragments from the \mbox{RePAIR} dataset~\cite{tsesmelisReassemblingRePAIRDataset2025} into granular media such that their painted surfaces are parallel and \SI{5}{mm} above the surface of the granular media.
\figref{fig:goal_examples} shows an examplary 3D render of an utilized fresco fragment, which was scanned by archaeologists.

On the real robotic system, we initially measure the mean height of the goal area and fit the goal shape onto the surface of the granular media.
This results in an adjusted goal height map that considers the unevenness of the media's initial surface.

Note that the larger the task, the completion time scales in proportion to the dimensions of the goal area.
Hence, while larger shapes are trainable, the chosen dimensions are a trade-off between training time and showcasing the agent's capabilities.

\subsubsection*{\textbf{Metrics}}
Based on the difference between the goal height map and the current height map at the end of 100 evaluation episodes, we calculate the absolute mean cell height difference~\(\hat{d}\) \textbf{(Height~Diff.)} within the goal area.
Furthermore, we determine the percentage of grid cells within the goal area that have been changed \textbf{(Changed)} by the EE. To calculate the number of steps that the approach requires to form the goal shape \textbf{(Execution)}, we count the steps from the start of the episode until the EE has left the granular medium for at least three consecutive steps.
For all metrics, we performed the Mann-Whitney U test~\cite{mannTestWhetherOne1947} to check for statistical significance between our and the other approaches.

%%%%%%%%%%%%%%%%%%%%%%%%
\subsection{Experimental Results}
\label{subsec:results}
\begin{table*}[h]
        \begingroup
        \setlength{\tabcolsep}{2pt}
        \centering
        \caption{Quantitative evaluation results.}
        \label{tab:evaluation}
            \begin{tabularx}{\linewidth}{X|PPPPP|PPP}
            \toprule
            \multicolumn{1}{c|}{\textbf{Metric}}
            & \multicolumn{2}{c}{\textbf{DELTA~(Ours)}} & \multicolumn{2}{c}{\textbf{PROG}} & \multicolumn{2}{c}{\textbf{NO-M}}
            & \multicolumn{2}{c}{\textbf{B-CPP}} & \multicolumn{2}{c|}{\textbf{RAND}}
            & \multicolumn{2}{c}{\textbf{DELTA}-\(H^\text{P}\)\phantom{\(^{\star\star}\)}} & \multicolumn{2}{c}{\textbf{PROG}-\(H^\text{P}\)\phantom{\(^{\star\star}\)}} & \multicolumn{2}{c}{\textbf{NO-M}-\(H^\text{P}\)\phantom{\(^{\star\star}\)}}
            \\
            \midrule
            a)~Height~Diff.~[mm]~\textdownarrow
              & \mathbf{3.4} & 1.1
              & 4.5 & 1.9^{\star\star\star}
              & 6.0 & 1.8^{\star\star\star}
              & 4.8 & 1.0^{\star\star\star}
              & 7.2 & 2.5^{\star\star\star}
              & 3.3 & 1.0
              & 4.5 & 1.8
              & 6.1 & 1.9
              \\
            b)~Changed~[\%]~\textuparrow
              & 97.4 & 10.4
              & 98.2 & 7.7
              & 3.4 & 12.8
              & \mathbf{100.0} & 0.0
              & 53.6 & 33.8
              & 95.5 & 14.9
              & 98.7 & 5.6
              & 1.0 & 6.7
              \\
            c)~Execution~[steps]~\textdownarrow
              & \mathbf{23.5} & 7.9
              & 34.9 & 7.6^{\star\star\star}
              & 38.1 & 7.0^{\star\star\star}
              & 44.0 & 16.8^{\star\star\star}
              & 27.2 & 14.0^{\star\star}
              & 22.3 & 7.8
              & 36.0 & 7.3
              & 39.3 & 4.1
              \\
            \bottomrule
            \end{tabularx}
        \endgroup
        \vspace{0.5ex}
        
        \noindent
        {%
        The metrics are based on the manipulated cells within the goal area during 100 evaluation episodes.
        On average DELTA achieves the significantly lowest mean value for the Height Diff. and Execution metric compared to PROG, NO-M, and both baselines.
        While B-CPP changes all goal cells, the trained RL agents relying on either one of the two reward formulations (DELTA and PROG) change more cells to the correct height.
        When removing the goal area movement reward (NO-M), the agent learns to completely avoid manipulations, which leads to the second highest Height Diff. value, lower than the random baseline (RAND), but higher than the B-CPP baseline.
        The agents relying on the reconstructed height map~\(H^\text{R}\) (left) successfully learn to cope with occlusions, as their results are close to the agents using the privileged height map~\(H^\text{P}\) (right). 
        \(\star\star\) and \(\star\star\star\) denote statistically significant results for our agent using the DELTA reward compared to all other approaches with \(p\)-value thresholds of \(0.01\) and \(0.001\) respectively.
        }
\end{table*}
\begin{table}[h]
	\begingroup
	\setlength{\tabcolsep}{2pt} % Default value: 6pt
	\centering
	\caption{Feature extractor ablation results.}
	\label{tab:evaluation_feature}
	\begin{tabularx}{\linewidth}{X|PPP}
            \toprule
            \multicolumn{1}{c|}{\textbf{Metric}} &
            \multicolumn{2}{c}{\textbf{DELTA~(Ours)}\phantom{\(^{\star}\)}} &
            \multicolumn{2}{c}{\textbf{ABL-CNN}\phantom{\(^{\star\star\star}\)}} &
            \multicolumn{2}{c}{\textbf{ABL-IMG}} \\
            \midrule
            a)~Height~Diff.~[mm]~\textdownarrow
              & \mathbf{3.4} & 1.1
              & 3.7 & 1.2
              & 4.6 & 1.7^{\star\star\star} \\
            b)~Changed~[\%]~\textuparrow
              & 97.4 & 10.4
              & \mathbf{97.7} & 6.2
              & 95.0 & 12.5 \\
            \bottomrule
	\end{tabularx}
	\endgroup
        \vspace{0.5ex}
        
        \noindent
        {%
        Our visual feature extractor (DELTA) shows the best performance for the Height Diff. metric compared to two ablated versions using CNN-based extractors without the gating mechanism (ABL-CNN) and relying on pure depth images (ABL-IMG) instead of the reconstructed height map~\(H^\text{R}\).
        \(\star\star\star\)~denotes statistically significant results for our feature extractor (DELTA) compared to all ablations with \(p\)-value thresholds of \(0.001\).
        }
\end{table}

The policies optimize the reward components discussed in \secref{subsec:reward}, iteratively refining its manipulation strategy to shape the granular medium into the desired configuration.
The results presented in the first column of \tabref{tab:evaluation}, show that our best policy (DELTA) achieves a remaining absolute mean cell height difference \(\hat{d}\) of up to \SI{3.4}{mm} over all goal area cells, which is close to the privileged setting, and outperforming two baselines.
This indicates that the trained policy exhibits proficient manipulation behaviors, given that the agent has to learn the dynamics of collapsing cells to reach the desired goal configuration.

% ==============================
\subsubsection*{\textbf{Qualitative Results}}
For a qualitative assessment, we visualize rollouts of the policy, displaying the resulting height maps in \figref{fig:rollouts}.
The rollouts show the robot forming a rectangle shape in simulation and in the real world.
Furthermore, \figref{fig:goal_examples} shows the qualitative results of several goal shapes ranging from a simple rectangular shape to a complex shape resulting from the negative of an archaeological fragment.

% ==============================
\subsubsection*{\textbf{Quantitative Results}}
\begin{figure}[b]
    \centering
    \includegraphics[width=\linewidth]{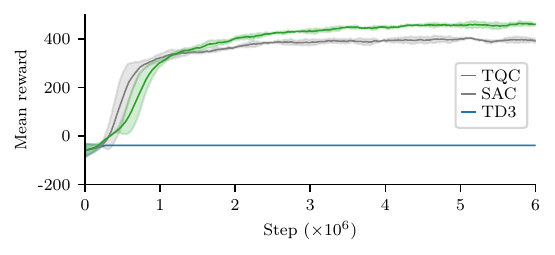}
    \vspace{-1.75\baselineskip}
    \caption{Using three different seeds, the RL training with TQC~\cite{kuznetsovControllingOverestimationBias2020} results in higher mean rewards compared to SAC~\cite{haarnojaSoftActorCriticOffPolicy2018a}, while TD3~\cite{fujimotoAddressingFunctionApproximation2018a} does not converge showing poor performance.}
    \label{fig:rl_algo_ablation}
\end{figure}
To understand the effectiveness of our proposed reward design, we compare the performance of our trained policy using the delta reward \textbf{(DELTA)} against the progressive reward \textbf{(PROG)} and the reward ablation where we remove the goal area movement reward \textbf{(NO-M)}.
First, removing the goal area movement reward leads to a policy that entirely avoids manipulation behaviors, since discovering strategies that shape the granular medium to match the desired configuration are challenging to discover without any guidance.
As a result, the policies perform no better than a random baseline \textbf{(RAND)}.
Second, even though the Boustrophedon Coverage Path Planning baseline \mbox{\textbf{(B-CPP)}} changes all goal cells, our approach has \SI{31}{\percent} higher accuracy in achieving the target heights than the B-CPP baseline.
Third, our DELTA approach is significantly faster in shaping the goal height maps.
Fourth, using the reconstructed height map~\(H^\text{R}\) in simulation leads to similar performance as relying on the privileged height map~\(H^\text{P}\), while only the former is deployable to the real world.
In summary, the results indicate that our full reward formulation accelerates learning and significantly improves the final performance compared to all other approaches.

% ==============================
\subsubsection*{\textbf{Feature Extractor Ablation}}
To analyze the effectiveness of our feature extractor described in \secref{subsec:architecture}, we compare its performance to two ablated versions: The first one \mbox{(\textbf{ABL-CNN})} does not contain the gating mechanism and uses a three-channel CNN for the height map and the two mask observations instead.
The second ablated version (\textbf{ABL-IMG}) also uses a CNN-based encoder, but instead of the reconstructed difference height map it directly uses the depth image from the camera together with the two mask observations.
The quantitative results in \tabref{tab:evaluation_feature} show that using our feature extractor with the delta reward achieves the best Height Diff. performance compared to both ablated versions, with a mean of \SI{3.4}{mm}.
Relying on pure depth images (ABL-IMG) prevents the trained agent from learning suitable features, such that it is not able to correctly lower the goal cells, resulting in a significantly higher mean height difference of \SI{4.6}{mm} compared to our feature extractor.

% ==============================
\subsubsection*{\textbf{RL Algorithm Ablation}}
Furthermore, we compare the performance of the TQC algorithm with two other commonly used off-policy RL algorithms, namely Soft Actor-Critic~(SAC)~\cite{haarnojaSoftActorCriticOffPolicy2018a} and Twin Delayed Deep Deterministic Policy Gradient (TD3)~\cite{fujimotoAddressingFunctionApproximation2018a}, using the same training parameters (see \tabref{tab:rl_training}) and three different seeds.
\figref{fig:rl_algo_ablation} shows the mean rewards with three different training seeds.
Note that the mean reward values are based on 25 evaluation episodes.
Training with TQC leads to a fast convergence and it outperforms SAC in terms of mean rewards.
Using TD3 does not converge to positive rewards, validating our design choice to utilize TQC.

% ==============================
\subsubsection*{\textbf{Real World Transfer}}
Finally, we zero-shot deployed the agent trained in simulation to the real robotic system.
\figref{fig:transfer} shows one qualitative episode with an exemplary rectangle shape.
The performance of our agent in the real world is similar to the one in simulation (compare \figref{fig:simulation} and \figref{fig:transfer}), which demonstrates that our agent can successfully be deployed on a real robot.

%%%%%%%%%%%%%%%%%%%%%%%%%%%%%%%%%%%%%%%%%%%%%%%%%%%%%%%%%%%%%%%%%%%%%%%%%%%%%%%%
\section{Conclusion}
\label{sec:conclusion}
Manipulating granular media requires interactive behaviors that adapt to changes in the medium’s state in a closed-loop fashion.
However, traditional modeling approaches require extensive engineering efforts to shape granular media due to its high-dimensional configuration space and its deformable nature.
Overcoming these shortcomings with reinforcement learning has remained a challenge.
In this work, we addressed this by developing suitable observation representations and reward functions that together enable a stable and efficient training.

Our approach significantly outperforms two baselines in terms of target shape accuracy, achieving the lowest difference between the desired and final medium configuration.
Furthermore, we demonstrated that policies trained entirely in simulation using depth-based observations can transfer zero-shot to real-world robotic systems, underscoring the applicability and robustness of our method.

\balance
\bibliographystyle{IEEEtran}
\bibliography{bibliography}
\end{document}